\documentclass[10pt,twocolumn,letterpaper]{article}

\usepackage{wacv}              

\definecolor{wacvblue}{rgb}{0.21,0.49,0.74}
\usepackage[pagebackref,breaklinks,colorlinks,allcolors=wacvblue]{hyperref}
\usepackage{booktabs}

\def\wacvPaperID{401} 
\def\confName{WACV}
\def\confYear{2027}

\title{PREEMPT: Proactive Robustness Evaluation under Editing for Media Protection Techniques}

\author{Qiuyu Tang\\
Lehigh University\\
Bethlehem, PA\\
{\tt\small qit220@lehigh.edu}
\and
Aparna Bharati\\
Lehigh University\\
Bethlehem, PA\\
{\tt\small apb220@lehigh.edu}
}

\begin{document}
\maketitle
\begin{abstract}
Proactive image protection methods aim to defend visual content against unauthorized manipulation and generative misuse by embedding protective signals into images before release. Evaluating whether these protections remain effective under image editing is therefore critical. However, existing evaluations remain largely limited to one-shot edits, providing only a static view of robustness while overlooking its evolution under realistic multi-step editing workflows. To systematically characterize robustness dynamics under sequential editing across various protection paradigms, we introduce PREEMPT, a holistic evaluation framework. It evaluates protection robustness under both one-shot and sequential editing scenarios, covering image transformations, semantic edits, homogeneous and heterogeneous edit compositions. To enable consistent analysis across protection paradigms with different underlying objectives, PREEMPT normalizes task-specific effectiveness into protection survival trajectories and summarizes long-horizon robustness through trajectory-based diagnostics. Using this framework, we reveal that degradation behaviors remain hidden under conventional one-shot evaluation. Furthermore, perturbation- and watermark-based protection methods exhibit different failure modes, and the first semantic edit largely determines subsequent degradation trajectories. PREEMPT provides a standardized protocol for studying long-horizon robustness and offers new insights into the degradation dynamics of proactive image protection.

\end{abstract}
    
\section{Introduction}
\label{sec:intro}

Recent advances in generative AI have made high-quality image editing increasingly accessible. Modern diffusion models~\cite{ho2020denoising, song2021denoising} enable realistic image manipulation~\cite{brooks2023instructpix2pix, couairon2022diffedit} and personalization~\cite{ruiz2023dreambooth, hu2022lora, gal2022image} from only a few user instructions, but these capabilities also raise concerns about unauthorized content modification, identity misuse, and style imitation. To mitigate such risks, proactive image protection methods~\cite{salman2023raising, jeong2025faceshield, tang2025styleprotect, wen2023treerings, fernandez2023stable} have been proposed to embed protective signals into images before release. Existing approaches mainly fall into two categories: perturbation-based methods~\cite{liu2024metacloak, shan2023glaze, xu2024perturbing}, which attempt to disrupt downstream generation behaviors, and watermark-based methods~\cite{liu2025watermarking, ye2024duaw, cui2025diffusionshield}, which enable ownership verification and content tracing after manipulation.


Existing robustness evaluations~\cite{zhang2018unreasonable, hessel2021clipscore, heusel2017gans, xu2025lmm4edit} suffer from three major limitations. First, experimental protocols differ substantially across studies, with inconsistent choices of datasets, editing models, and evaluation criteria. Second, robustness is typically evaluated under isolated one-shot edits, while realistic image manipulations often involve multiple sequential operations whose cumulative effects remain poorly understood. Third, evaluations are method-specific: perturbation- and watermark-based methods are designed for different objectives and consequently rely on different performance metrics. Together, these limitations hinder fair comparison across protection methods and provide only a limited understanding of robustness under realistic edits.

To address these limitations, we introduce the \textbf{Pr}oactive \textbf{E}valuation under \textbf{E}diting for \textbf{M}edia \textbf{P}rotection \textbf{T}echniques, (\textbf{PREEMPT}), a robustness evaluation framework for proactive image protection. PREEMPT systematically evaluates protection methods under both one-shot and long-horizon sequential editing scenarios, covering homogeneous semantic chains and heterogeneous mixed editing workflows. To enable consistent analysis across heterogeneous protection paradigms, PREEMPT further introduces a trajectory-based robustness representation consisting of Protection Survival Ratio (PSR), Protection Survival Curves, and Protection AUC, which respectively provide a unified robustness representation, trajectory-level visualization, and compact robustness summarization across heterogeneous protection paradigms.
Our contributions are:
\begin{itemize}
\itemsep0em
\item We propose PREEMPT, the first unified robustness evaluation framework for proactive image protection, which systematically evaluates protection methods under both one-shot and long-horizon sequential editing scenarios.

\item We introduce a trajectory-based robustness characterization based on the Protection Survival Curves, together with Protection Survival Ratio (PSR) and Protection AUC, enabling consistent robustness analysis across perturbation- and watermark-based protection paradigms.

\item Using PREEMPT, we uncover several previously overlooked properties of proactive image protection, including continuous robustness evolution under sequential editing, paradigm-specific evolution patterns, ranking instability beyond one-shot evaluation, and the dominant role of semantic edits in long-horizon robustness.

\end{itemize}

\section{Related Work}
\label{sec:related_work}
\subsection{Proactive Image Protection}

Proactive image protection aims to safeguard visual assets before they are released by embedding protective signals into images, thereby preventing unauthorized editing, personalization, or redistribution. Existing approaches can be broadly categorized into \emph{watermark-based} and \emph{perturbation-based} methods, which differ in both their protection objectives and underlying mechanisms. 

Watermark-based methods~\cite{lu2024robust, lei2024diffusetrace, zhang2024editguard, liu2025watermarking} embed identifiable information into images for ownership attribution, authenticity verification, and provenance tracing. Traditional watermarking techniques have recently been extended to diffusion-based generative models. Representative approaches include StegaStamp~\cite{Tancik_2020_CVPR}, which embeds recoverable signatures for copyright verification, Tree-Ring Watermarks~\cite{wen2023treerings}, which inject structured patterns into diffusion latents for provenance tracing, Stable Signature~\cite{fernandez2023stable}, which integrates watermark generation into image synthesis, and WAM~\cite{sander2025watermark}, which improves watermark robustness against common image editing and post-processing operations. Rather than preventing manipulation itself, these methods aim to preserve the recoverability of embedded messages after image modifications. 

In contrast, perturbation-based methods~\cite{wang2024edit, shan2023glaze, mi2024visual, xue2023toward} proactively interfere with downstream image editing and model customization by introducing carefully optimized adversarial perturbations. 
Representative perturbation-based methods differ substantially in their protection mechanisms. PhotoGuard~\cite{salman2023raising} disrupts the diffusion inference pipeline through adversarial perturbations, whereas Glaze~\cite{shan2023glaze} protects artistic styles by shifting feature representations toward surrogate targets. To disrupt the generation quality, AdvDM~\cite{liang2023adversarial} further formulates perturbation generation directly using the diffusion objective, MIST~\cite{liang2023mist} improves perturbation transferability via semantic-aware adversarial optimization. For personalization mimicry, Anti-DreamBooth~\cite{van2023anti} explicitly optimizes against the dynamic fine-tuning process.
Unlike watermarking, perturbation-based methods seek to reduce the fidelity of downstream generation while remaining visually imperceptible. 

Although the two paradigms pursue different protection objectives~\cite{tang2025watermarks, tang2025perturbation}, both fundamentally rely on the persistence of embedded protective signals throughout editing operations. Their robustness is therefore determined not only by the protection algorithm itself but also by the editing process that follows image release. This shared dependency motivates the need for a unified robustness evaluation framework capable of consistently assessing heterogeneous protection paradigms under realistic editing workflows.



\subsection{Existing Robustness Evaluation}

Existing robustness evaluation protocols differ across protection paradigms. Watermark-based methods~\cite{rabbani2024waves} typically assess robustness by measuring message retrieval accuracy or decoding success rate after image manipulations, with the objective of preserving watermark detectability throughout the editing process. Consequently, robustness is commonly evaluated under a collection of image transformations or semantic editing attacks by examining whether the embedded message can still be recovered~\cite{lu2024robust}. 
Perturbation-based methods usually evaluate two complementary aspects: perturbation imperceptibility and protection effectiveness. Imperceptibility is measured using perceptual image quality metrics such as LPIPS~\cite{zhang2018unreasonable}, while protection effectiveness is typically quantified by applying a single semantic edit and measuring the semantic similarity between the edited result and the protected image.

Besides semantic edits, robustness against image transformations has become a standard component of protection evaluation. Common operations include JPEG compression, Gaussian blur, crop-and-resize, horizontal flipping, and brightness/contrast adjustment, which are widely adopted to simulate the automatic preprocessing pipelines applied by online social networks (OSNs) after image upload~\cite{liu2023generating}. Since adversarial perturbations are particularly vulnerable to these lossy transformations, many recent protection methods explicitly evaluate robustness under such operations to verify that the embedded perturbations remain effective after post-processing. 

Despite adopting different objectives and evaluation metrics, existing protocols share several common limitations. 
First, robustness is predominantly evaluated under isolated one-shot manipulations. 
In contrast, real-world image editing is an iterative process. It typically involves sequential operations, continuous refinement, and history navigation~\cite{10.1145/2856400.2856417, chen2025edival, li2025editthinker}. Consequently, long-horizon robustness under sequential editing remains largely unexplored.
Second, evaluations rely primarily on endpoint measurements after editing and therefore fail to characterize how protection robustness evolves throughout the editing process. Finally, the use of method-specific metrics prevents direct comparison across perturbation- and watermark-based protection paradigms. 
These limitations motivate the need for a unified robustness evaluation framework that not only enables fair comparison across heterogeneous protection paradigms but also characterizes robustness as an evolving process rather than a single endpoint.
\section{PREEMPT}
\label{sec:method}

\subsection{Evaluation Pipeline}

\begin{figure*}[ht]
  \centering
    \includegraphics[width=\linewidth]{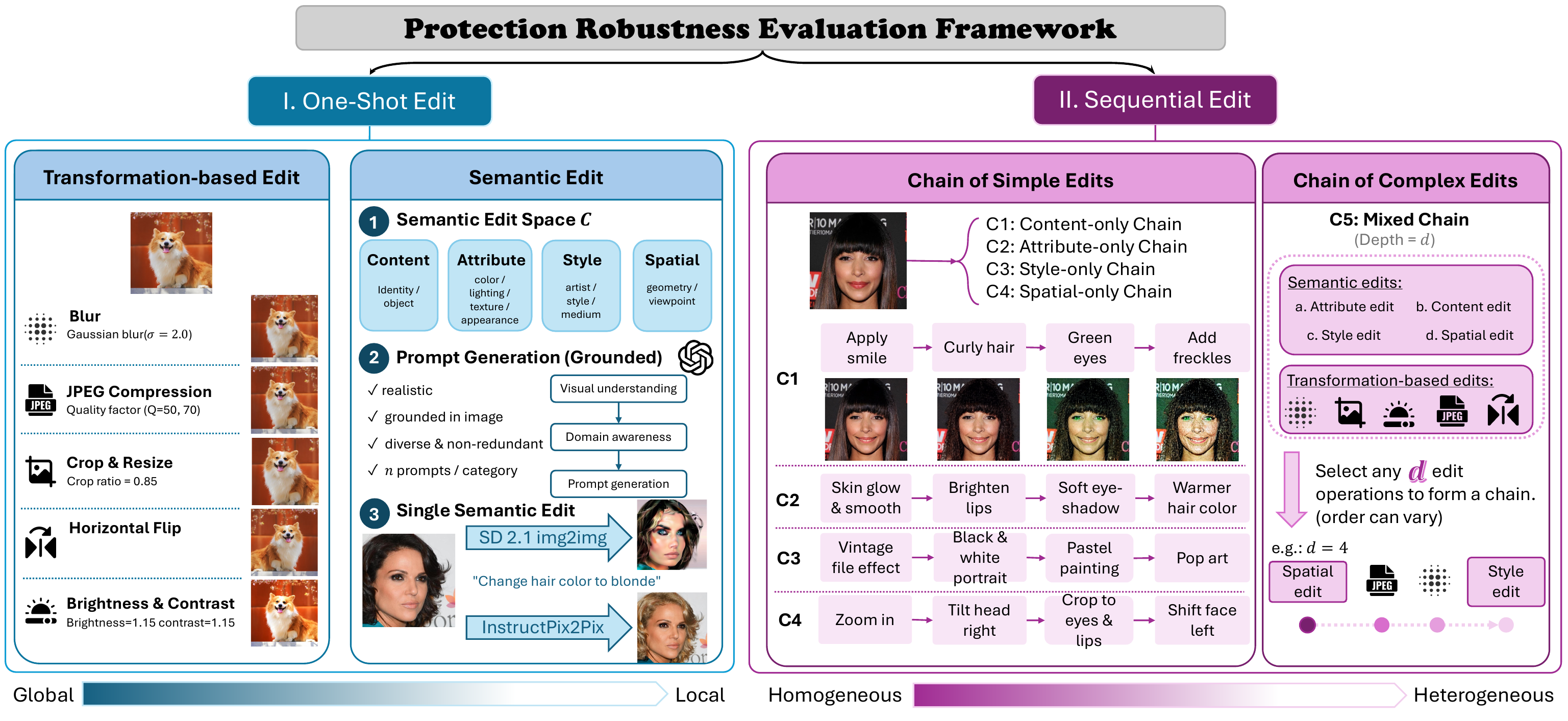}
    \vspace{-20pt}
    \caption{Overview of PREEMPT. Our framework evaluates proactive image protection methods under both one-shot and sequential editing scenarios. One-shot evaluation includes image-transformation-based edits and semantic edits, while sequential evaluation considers both homogeneous and heterogeneous edit chains to simulate realistic multi-step manipulations.}
    \label{fig:overview} 
    \vspace{-10pt}
\end{figure*}

To comprehensively simulate realistic post-deployment manipulations, we design a two-branch evaluation framework (see Fig.~\ref{fig:overview}) that evaluates protection robustness under both one-shot and long-horizon editing scenarios.  

\noindent\textbf{One-Shot Edit.} 
PREEMPT first evaluates robustness under isolated editing operations. One-shot evaluation consists of two complementary categories: \emph{Image-Transformation-based} and \emph{Semantic Edits}. Image Transformations simulate common pixel-level image processing operations encountered after image release, whereas semantic edits  evaluate robustness against content-level modifications generated by modern image editing models.

Transformation-based edits include commonly used image manipulations such as Gaussian blur, JPEG compression, crop-and-resize, horizontal flipping, and brightness/contrast adjustment. 
Semantic edits are organized into four categories: Content, Attribute, Style, and Spatial, covering modifications to semantic content, visual attributes, artistic style, and spatial configuration, respectively. Category-specific editing prompts are automatically generated by Vision-Language Models and executed by state-of-the-art image editing models to produce one-shot semantic manipulations.
Together, transformation-based and semantic edits provide complementary evaluations of protection robustness under isolated editing scenarios.

\noindent\textbf{Sequential Edit.}
PREEMPT further evaluates robustness under long-horizon editing workflows. Sequential edit pipelines are composed of multiple editing operation combinations. From homogeneous to heterogeneous settings, we consider two complementary settings: \emph{Simple Chains} and \emph{Mixed Chains}.
Simple Chains consist of homogeneous semantic edits from a single semantic category, including content-only (C1), attribute-only (C2), style-only (C3), and spatial-only (C4). These chains isolate robustness degradation within individual semantic dimensions. 
Mixed Chains combine semantic edits with image-transformation operations to simulate practical editing workflows. Given a chain depth $d$, edit operations are sequentially sampled from the editing pool to generate diverse editing trajectories.

\begin{figure*}[th]
    \centering
    \includegraphics[width=\linewidth]{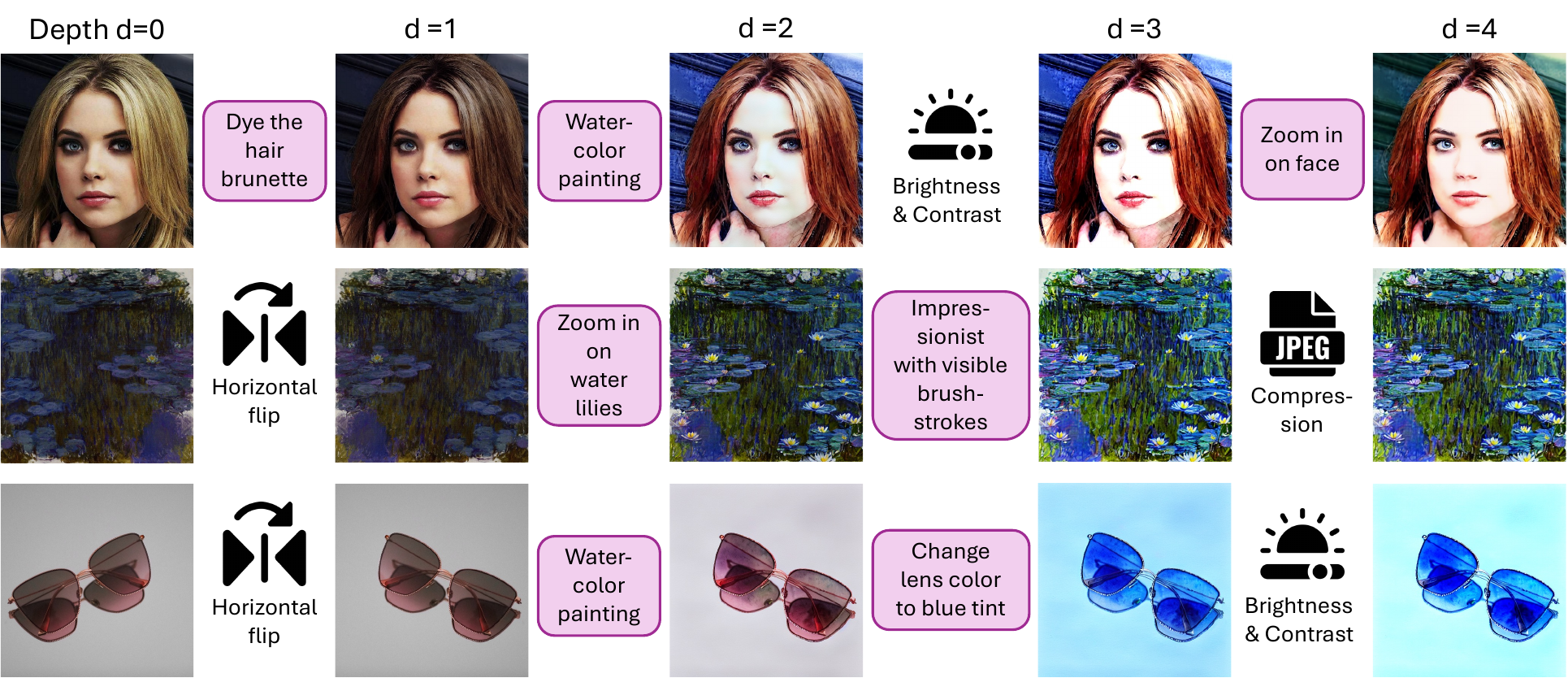}
    \vspace{-15pt}
    \caption{Representative heterogeneous editing trajectories. Rows correspond to faces, artworks, and objects. Each trajectory consists of alternating transformation-based and semantic edits, illustrating the mixed editing workflow used in our sequential robustness evaluation.}
    \label{fig:qualitative}
    \vspace{-10pt}
\end{figure*}

Figure~\ref{fig:qualitative} presents representative heterogeneous editing trajectories. Each trajectory combines transformation-based and semantic edits to simulate realistic editing workflows, motivating the need for a unified representation of robustness evolution. 

\subsection{Protection Survival Representation}

Existing proactive protection paradigms rely on different robustness metrics. Perturbation-based methods typically measure semantic similarity between the edited and original images, whereas watermark-based methods evaluate message retrieval accuracy. Since these paradigm-specific metrics differ in both scale and interpretation, directly comparing robustness across protection methods is difficult.

To quantify robustness evolution in a unified manner, we define the normalized \emph{Protection Survival Ratio (PSR)} as:
{\setlength{\abovedisplayskip}{2pt}
\setlength{\belowdisplayskip}{2pt}
\begin{equation}
\tilde{S}(d)=\frac{S(d)}{S(0)},
\label{eq:psr}
\end{equation}}

where $S(d)$ denotes the robustness metric adopted by the protection paradigm after editing depth $d$, and $S(0)$ is the metric value of the protected-but-unedited image. For perturbation-based methods, $S(d)$ corresponds to semantic similarity, while for watermark-based methods it corresponds to message retrieval accuracy.
By normalizing with respect to the initial protected state, PSR removes differences in metric scales and expresses robustness as a relative survival trajectory over editing depth. This formulation enables robustness evolution to be analyzed consistently across heterogeneous protection paradigms while preserving the interpretation of each underlying evaluation metric.

Notably, PSR is designed to characterize the evolution of robustness rather than to unify the underlying task-specific metrics themselves. Consequently, trajectory-based analyses, such as survival curves and Protection AUC, can be applied consistently to different protection paradigms without altering their original evaluation objectives.
For perturbation-based methods under semantic edits, low similarity after editing can reflect either the intended semantic change or an additional effect of protection.
To disentangle the two, we introduce a matched \emph{clean-trajectory control}: unprotected images (``None'') undergo the same editing chains as protected images, and we report
{\setlength{\abovedisplayskip}{2pt}
\setlength{\belowdisplayskip}{2pt}
\begin{equation}
\Delta(d)=S_{\mathrm{clean}}(d)-S_{\mathrm{prot}}(d),
\end{equation}}
where \(S_{\mathrm{clean}}(d)\) and \(S_{\mathrm{prot}}(d)\) are the paradigm-specific semantic similarities at depth \(d\).
A positive \(\Delta(d)\) indicates an additional protection-induced gap relative to the same clean editing path.
Watermarking methods are evaluated with bit accuracy and are excluded from this control, as they do not share the same semantic-similarity confound.

\subsection{Robustness Characterization}
To characterize long-horizon robustness, PREEMPT provides both a visual representation and a compact summary of protection survival trajectories.

\noindent\textbf{Protection Survival Curve.}
Given the normalized PSR, robustness under sequential editing can be represented as a protection survival curve, $\{\tilde{S}(0), \tilde{S}(1), \ldots, \tilde{S}(D)\}, $ where each point corresponds to the normalized protection effectiveness after editing depth $d$. Starting from the normalized initial state ($S(0)=1$), the resulting curve characterizes how protection effectiveness evolves throughout sequential editing. The curve provides an intuitive representation of robustness evolution and facilitates comparison across protection methods and editing scenarios.

\noindent\textbf{Protection AUC.}
While survival curves provide an intuitive visualization of robustness evolution, comparing massive trajectories across heterogeneous methods and scenarios can be cumbersome. We therefore summarize each trajectory using the area under the protection survival curve:
\vspace{-10pt}
\begin{equation}
\mathrm{AUC}
=
\frac{1}{D}
\sum_{d=0}^{D-1}
\frac{\tilde{S}(d)+\tilde{S}(d+1)}{2},
\end{equation}
\vspace{-10pt}

where $D$ denotes the maximum editing depth. Since $\tilde S(d)\in[0,1]$, the resulting AUC is also normalized to $[0,1]$, with larger values indicating stronger overall robustness throughout the editing process.

\section{One-shot Robustness}
\label{sec:one_shot_results}
\subsection{Experimental Setup}

To demonstrate PREEMPT, we instantiate the framework using representative proactive image protection methods from both perturbation-based and watermark-based paradigms. Specifically, we evaluate four perturbation-based methods (PhotoGuard~\cite{salman2023raising}, Mist~\cite{liang2023mist}, AdvDM~\cite{liang2023adversarial}, and SDST~\cite{xue2023toward}) together with four recent watermarking methods (Robust-Wide~\cite{hu2025robust}, TrustMark~\cite{Trustmark-ICCV-2025}, VINE~\cite{lu2024robust} and WAM~\cite{sander2025watermark}). These methods are selected to demonstrate the applicability of PREEMPT across various protection paradigms. To ensure fair comparison, all perturbation-based methods are generated using a standard perturbation budget ($\epsilon=12/255$) and 100 optimization steps. Clean images serve as the no-protection baseline (\textit{None}).

Protection effectiveness is evaluated using the task-specific metrics corresponding to each protection paradigm. For perturbation-based methods, semantic similarity is measured using domain-specific feature representations: ArcFace~\cite{deng2019arcface} identity similarity for biometric faces, CSD~\cite{somepalli2024measuring} style embedding similarity for artworks, and DINOv2~\cite{oquab2023dinov2} feature similarity for natural scene images. The resulting scores are normalized to a common scale for cross-domain comparison. For watermark-based methods, robustness is measured by message retrieval bit accuracy using each method's official encoding and decoding pipeline. 
Following the original implementations, we embed the message ``Watermarked'' for VINE and ``WAMK'' for WAM according to their payload capacities, a fixed 64-bit binary payload for Robust-Wide, and a fixed 100-bit binary payload for TrustMark (ECC disabled).

Image-transformation-based edits include Gaussian blur, JPEG compression, crop-and-resize, horizontal flipping, and brightness/contrast adjustment. These transformations~\cite{verdoliva2020media} represent typical post-processing encountered after image release. For semantic edits, four editing prompts are generated for each semantic category using GPT-4.1-mini~\cite{openai2023gpt4} and executed by two representative image editing models, Stable Diffusion v2.1~\cite{rombach2022high} and InstructPix2Pix~\cite{brooks2023instructpix2pix}. Since PREEMPT is independent of the underlying editing model, alternative editors can be incorporated without modifying the evaluation protocol.

The evaluation benchmark covers three representative visual domains: artwork images from WikiArt~\cite{artgan2018}, biometric face images from CelebA-HQ~\cite{liu2015deep, karras2018progressive}, and object-centric images from DreamBooth~\cite{ruiz2023dreambooth}. Curated subsets from these datasets are used throughout all experiments. Complete implementation details, parameter settings, and prompt templates are provided in the Supplementary Material (SM).

\subsection{Image-Transformation-based edits}

Image-transformation-based edits evaluate protection robustness under common post-processing operations encountered during image sharing. The quantitative results for perturbation- and watermark-based methods are reported in Tab.~\ref{tab:per_l1_semantic_similarity} and Tab.~\ref{tab:wm_l1_retrieval_accuracy}, respectively. We also report the ``No edit'' column as the no-transformation anchor (protected vs.\ clean before any transform).


\begin{table}[t]
\centering
\caption{Effectiveness of perturbation protection methods under one-shot image transformation.
Metric: semantic similarity ($\downarrow$).}
\label{tab:per_l1_semantic_similarity}
\vspace{-5pt}
\resizebox{\columnwidth}{!}{
\begin{tabular}{lccccccc|c}
\toprule
Method & No edit & JPEG$_{70}$ & JPEG$_{50}$ & Bright.\ \& Contr. & Blur & Crop & Flip & Avg. \\
\midrule
\textit{None} & 1.000 & 0.988 & 0.975 & 0.976 & 0.971 & 0.940 & 0.966 & 0.969 \\
AdvDM & 0.872 & 0.874 & 0.872 & 0.869 & 0.880 & 0.824 & 0.858 & 0.863 \\
MIST & 0.901 & 0.895 & 0.889 & 0.890 & 0.881 & 0.855 & 0.884 & 0.882 \\
PhotoGuard & 0.900 & 0.895 & 0.890 & 0.890 & 0.880 & 0.856 & 0.885 & 0.883 \\
SDST & 0.909 & 0.899 & 0.896 & 0.896 & 0.884 & 0.865 & 0.890 & 0.888 \\
\bottomrule
\end{tabular}
}
\vspace{-5pt}
\end{table}



\begin{table}[t]
\centering
\caption{Effectiveness of watermarking methods under one-shot image transformations. Metric: message retrieval accuracy ($\uparrow$).}
\label{tab:wm_l1_retrieval_accuracy}
\vspace{-5pt}
\resizebox{\columnwidth}{!}{
\begin{tabular}{lccccccc|c}
\toprule
Method & No edit & JPEG$_{70}$ & JPEG$_{50}$ & Bright.\ \& Contr. & Blur & Crop & Flip & Avg. \\
\midrule
Robust-Wide & 1.000 & 1.000 & 1.000 & 0.987 & 1.000 & 0.539 & 0.452 & 0.830 \\ 
TrustMark & 1.000 & 0.999 & 0.998 & 0.982 & 1.000 & 1.000 & 0.999 & 0.996 \\ 
VINE & 1.000 & 1.000 & 1.000 & 0.998 & 0.999 & 0.504 & 0.465 & 0.828 \\
WAM & 1.000 & 0.998 & 0.994 & 1.000 & 1.000 & 1.000 & 1.000 & 0.999 \\

\bottomrule
\end{tabular}
}
\vspace{-18pt}
\end{table}


Both protection paradigms remain largely robust under common image transformations. Compared with the unprotected baseline, perturbation-based methods consistently reduce semantic similarity across all evaluated transformations, while watermark-based methods maintain nearly perfect message retrieval accuracy under most operations. Crop-and-resize produces the largest performance degradation for both paradigms, whereas JPEG compression, Gaussian blur, and brightness/contrast adjustment have comparatively minor effects. These results indicate that routine post-processing alone is generally insufficient to invalidate existing proactive image protection methods.

\subsection{Semantic Edit}

Unlike image transformations, semantic edits directly modify image content and therefore provide a more challenging evaluation scenario. Unless otherwise specified, we report results obtained using InstructPix2Pix~\cite{brooks2023instructpix2pix} in the main paper, while the corresponding results using Stable Diffusion v2.1~\cite{rombach2022high} are provided in SM. The quantitative results are summarized in Tab.~\ref{tab:per_l2_semantic_similarity} and Tab.~\ref{tab:wm_l2_retrieval_accuracy}.

\begin{table}[t]
\centering
\caption{Protection effectiveness of perturbation-based methods under one-shot semantic edits. Metric: semantic similarity ($\downarrow$).}
\label{tab:per_l2_semantic_similarity}
\vspace{-8pt}
\resizebox{\columnwidth}{!}{
\begin{tabular}{lccccc|c}
\toprule

Method & No edit & Content & Attribute & Style & Spatial & Avg. \\
\midrule
\textit{None} & 1.000 & 0.760 & 0.720 & 0.740 & 0.695 & 0.729 \\
AdvDM & 0.872 & 0.267 & 0.234 & 0.331 & 0.249 & 0.270 \\
MIST & 0.901 & 0.505 & 0.384 & 0.495 & 0.492 & 0.469 \\
PhotoGuard & 0.900 & 0.490 & 0.372 & 0.486 & 0.462 & 0.453 \\
SDST & 0.909 & 0.458 & 0.376 & 0.508 & 0.495 & 0.459 \\
\bottomrule
\end{tabular}
}
\vspace{-10pt}
\end{table}

\begin{table}[t]
\centering
\caption{Protection effectiveness of watermarking methods under one-shot semantic edits. Metric: message retrieval accuracy ($\uparrow$). }
\label{tab:wm_l2_retrieval_accuracy}
\vspace{-8pt}
\resizebox{\columnwidth}{!}{
\begin{tabular}{lccccc|c}
\toprule
Method & No edit & Content & Attribute & Style & Spatial & Average \\
\midrule
Robust-Wide & 1.000 & 0.990 & 0.970 & 0.950 & 0.975 & 0.971 \\ 
TrustMark & 1.000 & 0.899 & 0.866 & 0.859 & 0.852 & 0.869 \\ 
VINE & 1.000 & 0.945 & 0.899 & 0.850 & 0.917 & 0.903 \\
WAM  & 1.000 & 0.894 & 0.805 & 0.733 & 0.837 & 0.817 \\
\bottomrule
\end{tabular}
}
\vspace{-18pt}
\end{table}

Compared with image transformations, semantic edits pose a substantially greater challenge to proactive image protection. Perturbation-based methods exhibit markedly lower protection effectiveness after a single semantic manipulation, while watermark-based methods also experience measurable retrieval degradation despite maintaining relatively high decoding accuracy. Across both protection paradigms, content-level modifications have a considerably stronger impact than low-level image transformations.

Nevertheless, these one-shot evaluations only measure protection effectiveness after an isolated edit and therefore provide static endpoint measurements. Whether such observations remain valid under realistic multi-step editing workflows is investigated in the next section.
\section{Long-Horizon Robustness}
\label{sec:sequential_result}

\subsection{Experimental Setup}
For long-horizon evaluation, we construct sequential editing chains with a default depth $d=4$ throughout the main paper. Results of Stable Diffusion v2.1, along with additional long-horizon experiments including sensitivity to chain length and edit ordering, are presented in SM.

We consider two complementary sequential settings. Simple chains repeatedly apply semantic edits from a single category, enabling controlled evaluation of robustness evolution within individual semantic dimensions. Complex chains consist of heterogeneous edit sequences by combining semantic edits and image transformations, simulating practical multi-step editing workflows. Unless otherwise specified, edit operations are randomly sampled using a fixed seed of 42 to ensure reproducibility.

We do not consider transformation-only chains. Consecutive image transformations primarily accumulate low-level distortions and can often be approximated by a single transformation with adjusted parameters. In contrast, sequential semantic edits progressively reconstruct image content, resulting in substantially different robustness evolution. We therefore focus homogeneous chains on semantic edits while using heterogeneous chains to approximate realistic editing workflows.

\subsection{Robustness Evolution}

\begin{figure*}[ht]
    \centering
    \includegraphics[width=.95\linewidth]{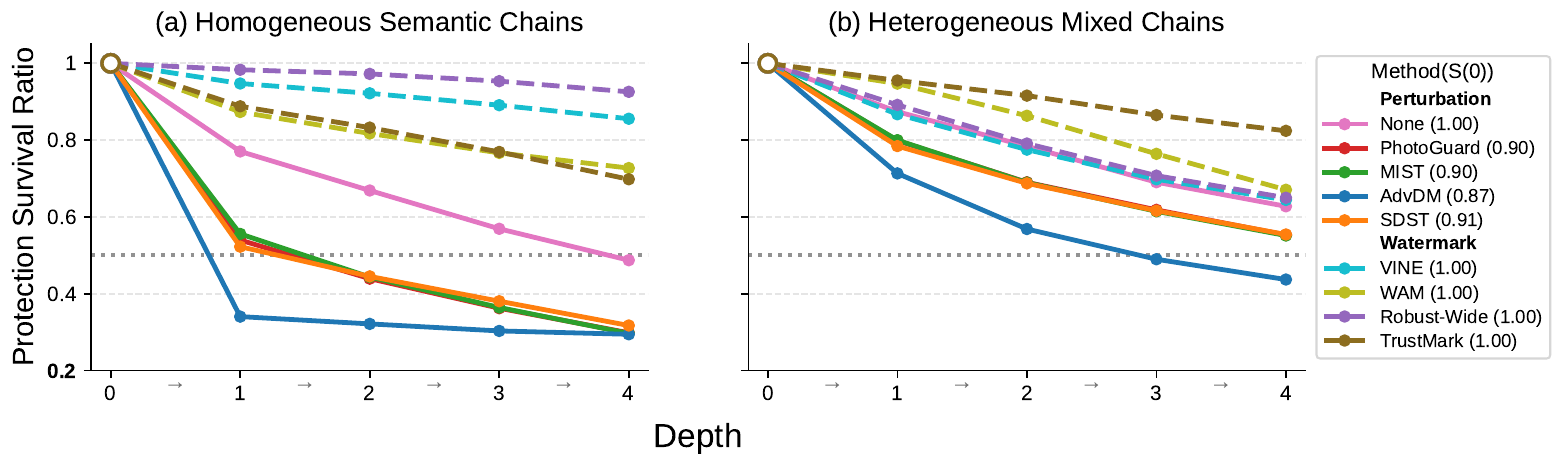}
    \vspace{-8pt}
    \caption{Protection survival curves under homogeneous semantic chains (left) and heterogeneous mixed chains (right). Robustness evolves progressively with editing depth rather than remaining a static endpoint. The x-axis denotes the editing depth ($D=4$), and the y-axis shows the normalized Protection Survival Ratio (PSR). Solid lines indicate perturbation-based methods, and dashed lines indicate watermark-based methods. The legend reports each method together with its pre-edit effectiveness, $S(0)$.}
    \label{fig:protection_survival_curve}
    \vspace{-10pt}
\end{figure*}

Figure~\ref{fig:protection_survival_curve} visualizes the evolution of protection robustness under sequential editing using the proposed PSR. Homogeneous semantic chains and heterogeneous mixed chains are shown separately to distinguish controlled semantic degradation from practical multi-step editing workflows.
Across both sequential settings, robustness evolves continuously with editing depth rather than collapsing after a single manipulation. All protection methods exhibit progressively decreasing PSR as additional edits are applied, indicating that robustness should be regarded as a dynamic property instead of a static endpoint.
Under the matched clean trajectory, unprotected images already exhibit substantial semantic drift---e.g., raw concept similarity drops to \(S(4)\approx0.49\) on homogeneous chains and \(S(4)\approx0.64\) on heterogeneous chains showing that sequential semantic editing itself reduces the measured signal.
Relative to this baseline, the evaluated perturbation methods consistently yield \(\Delta(d)>0\). Protection pushes similarity down by an additional gap beyond the intended edit
(see SM for clean-trajectory curves and $\Delta$ tables).

This observation highlights an important limitation of conventional one-shot evaluation. Although measuring protection effectiveness after a single edit provides a useful snapshot of robustness, it does not reveal how protection evolves under realistic multi-step editing. Our results show that protection degrades progressively as edits accumulate, motivating trajectory-based analyses that capture robustness dynamics over the entire editing process.

\subsection{Paradigm-specific Evolution}
\begin{figure}[htb]
\vspace{-5pt}
    \centering
    \includegraphics[width=.85\linewidth]{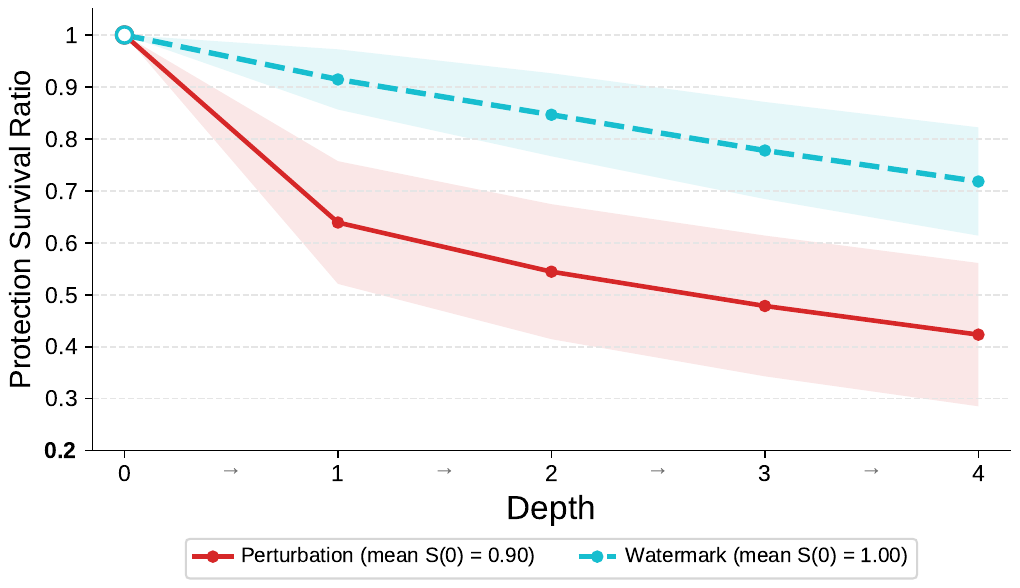}
    \vspace{-5pt}
    \caption{Protection survival curves aggregated by protection paradigms. Shaded regions indicate the standard deviation across methods within each protection paradigm. Perturbation- and watermark-based methods exhibit distinct robustness evolution patterns under sequential editing.}
    \label{fig:protection_family_survival_curve}
    \vspace{-15pt}
\end{figure}

Figure~\ref{fig:protection_family_survival_curve} compares the normalized robustness trajectories of perturbation- and watermark-based methods.Although both paradigms exhibit continuous robustness evolution under sequential editing, their evolution patterns differ substantially. Perturbation-based methods show a rapid change immediately after the first edit, followed by progressively diverging trajectories across methods. In contrast, watermark-based methods exhibit smoother and more consistent robustness evolution throughout the editing process. Sequential editing also increases the variability of robustness trajectories among perturbation-based methods, as reflected by the progressively wider spread across methods.

The observed difference is consistent with the different design objectives of the two protection paradigms. Perturbation-based methods protect images by disrupting downstream image generation, making their robustness closely coupled with semantic reconstruction during editing. Consequently, repeated semantic edits produce increasingly diverse robustness trajectories across different perturbation methods. In contrast, watermark-based methods aim to preserve an embedded signal for subsequent decoding. As long as sufficient watermark information survives each editing step, message retrieval remains relatively stable, leading to smoother robustness trajectories. These results indicate that robustness evolution is paradigm-dependent. Rather than enforcing a single robustness metric, PREEMPT accommodates paradigm-specific measurements within a unified trajectory-based evaluation framework, enabling consistent analysis and comparison of robustness evolution across different protection paradigms.



\subsection{Ranking Stability}

\begin{figure}[htb]
    \vspace{-5pt}
    \centering
    \includegraphics[width=.8\linewidth]{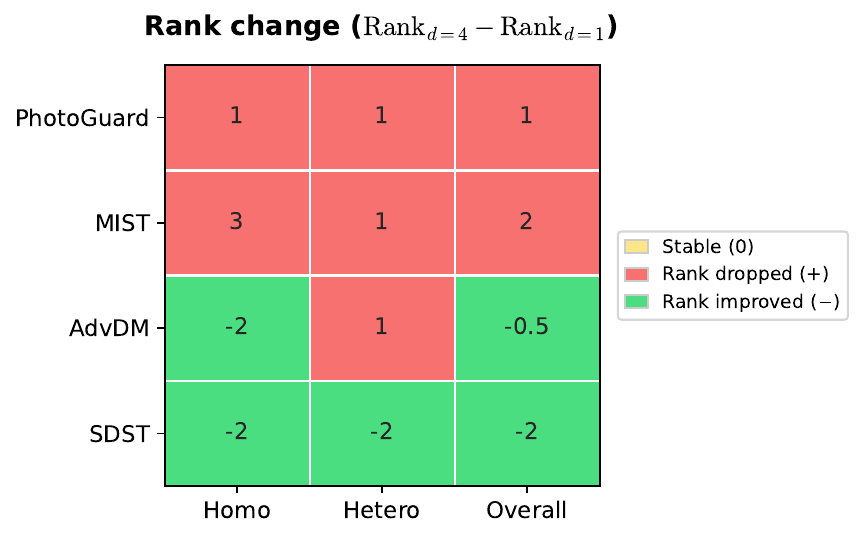}
    \vspace{-8pt}
    \caption{
    Average rank change from one-shot ($d=1$) to long-horizon ($d=4$) evaluation under homogeneous and heterogeneous edit chains. Positive (negative) values indicate worse (better) rankings after sequential editing, showing that one-shot rankings are not consistently preserved.}
    \label{fig:rank_shift_heatmap}
    \vspace{-18pt}
\end{figure}

Ranking instability is primarily observed for perturbation-based methods. We focus the analysis on perturbation-based methods, where ranking changes are more informative.
Figure~\ref{fig:rank_shift_heatmap} summarizes average ranking change from one-shot ($d=1$) to long-horizon ($d=4$) evaluation. Positive values indicate that a method drops in rank after sequential editing, whereas negative values indicate an improved ranking. Although all methods are evaluated under the same protocol, their rankings are not consistently preserved. For example, MIST, which ranks among the strongest methods after a single edit, consistently loses ranking under sequential evaluation, whereas AdvDM and SDST improve their ranks across multiple editing settings.

These observations suggest that one-shot robustness is not always predictive of long-horizon robustness. 
To quantify this gap beyond ranking changes, we test whether one-shot semantic robustness predicts long-horizon survival.
For each protection method, we take the mean one-shot concept similarity over the four semantic edit types as the predictor, and use the normalized protection survival AUC (PAUC) under sequential editing as the target.
Fitting a linear regression at the method$\times$chain level yields only a moderate association ($r{\approx}0.40$, $R^{2}{\approx}0.16$), indicating that one-shot scores explain little of the variance in long-horizon robustness.
Thus, endpoint measurements after a single edit are insufficient to characterize protection under multi-step editing; trajectory-based evaluation remains necessary.
Detailed scatter plots and correlation statistics are provided in SM. 
Methods exhibiting similar performance after a single edit may follow substantially different robustness trajectories under continued editing, eventually leading to different relative rankings. Consequently, evaluating proactive protection methods solely under one-shot settings may produce method rankings that do not generalize to realistic multi-step editing scenarios. 
Sequential evaluation therefore provides complementary information that cannot be inferred from one-shot robustness alone.




\subsection{The Role of the First Semantic Edit}

Having established that protection robustness evolves throughout sequential editing, we next investigate what drives this evolution. Since heterogeneous edit chains consist of both image transformations and semantic edits, we group editing trajectories according to whether the first operation is a physical transformation or a semantic edit. 

\begin{figure}[ht]
    \centering
    \includegraphics[width=\linewidth]{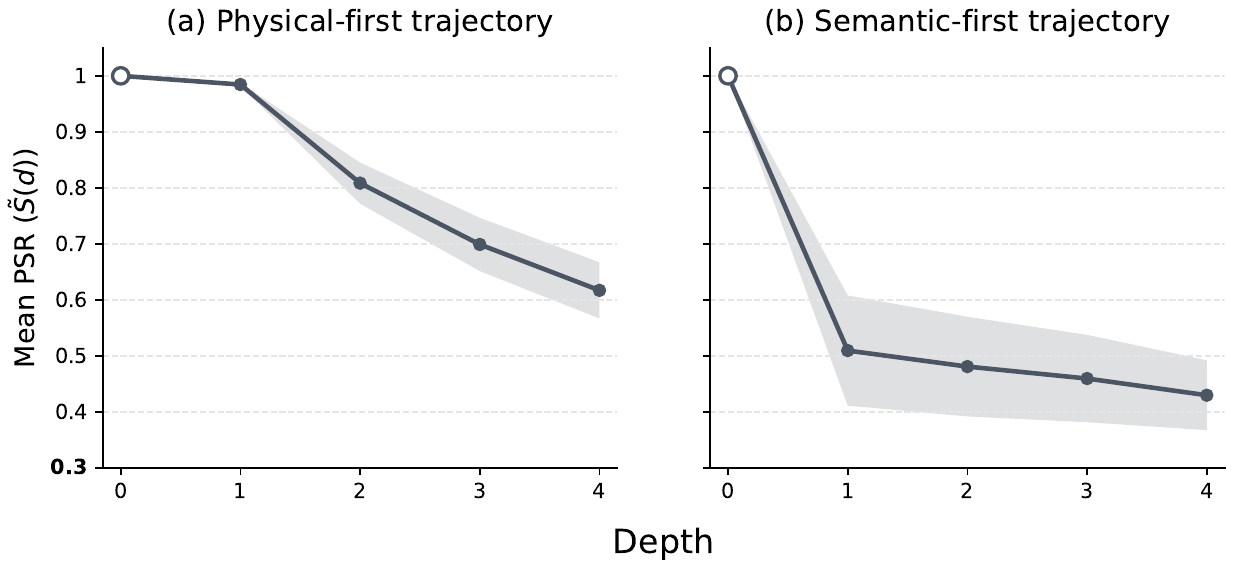}
    \vspace{-20pt}
    \caption{Robustness trajectories under heterogeneous edit chains grouped by the type of the first edit.
(a) Physical-first trajectories remain largely unchanged until the first semantic edit occurs.
(b) Semantic-first trajectories exhibit an immediate robustness transition followed by slower evolution.
Shaded regions denote the standard deviation across perturbation-based methods.
}
    \label{fig:first_edit_type}
    \vspace{-15pt}
\end{figure}

Figure~\ref{fig:first_edit_type} compares the resulting robustness trajectories. When the first operation is an image transformation, protection robustness remains nearly unchanged after the first editing step, and the major robustness transition occurs only after subsequent semantic edits are introduced (see Fig.~\ref{fig:first_edit_type} (a)). In contrast, when the first operation is semantic, robustness exhibits an immediate transition after the first editing step, followed by substantially slower evolution throughout the remaining editing process (see Fig.~\ref{fig:first_edit_type} (b)). This contrast suggests that robustness evolution is more strongly associated with semantic edits than with low-level image transformations. Unlike image transformations, which mainly alter low-level image statistics while preserving semantic content, semantic edits explicitly reconstruct the image content, directly affecting the information that proactive protection methods seek to preserve or disrupt.

We further define the first-step contribution: 
{\setlength{\abovedisplayskip}{2pt}
\setlength{\belowdisplayskip}{2pt}
\[ \alpha= \frac{\tilde{S}(0)-\tilde{S}(1)} {\tilde{S}(0)-\tilde{S}(D)}, \] }
which measures the fraction of the total robustness degradation occurring after the first editing step. The results are summarized in Tab.~\ref{tab:first-step-alpha}. Across homogeneous semantic chains, the first semantic edit accounts for approximately 63\%--93\% of the total degradation, indicating that most robustness change is concentrated in the initial semantic modification. In heterogeneous edit chains, however, this proportion consistently decreases to 44\%--52\%, suggesting robustness degrades more gradually across successive editing operations rather than being dominated by the first edit.

Taken together, these results indicate that semantic edits serve as the primary transition governing perturbation robustness, whereas image transformations largely preserve the existing robustness state until semantic reconstruction occurs. More importantly, they demonstrate that, under the evaluated editing scenarios, the evolution of protection robustness depends not only on editing depth but also on the composition of the editing sequence.

\begin{table}[t]
  \centering
  \caption{Contribution of the first editing step to the overall robustness evolution.
Higher $\alpha$ indicates that a larger proportion of robustness change occurs immediately after the first edit.
}
\vspace{-5pt}
  \label{tab:first-step-alpha}
  \begin{tabular}{lcc}
    \toprule
    Method & Homo. $\alpha$ & Hetero. $\alpha$ \\
    \midrule
    PhotoGuard & 0.65 & 0.45 \\
    MIST       & 0.63 & 0.45 \\
    AdvDM      & 0.93 & 0.52 \\
    SDST       & 0.69 & 0.44 \\
    \bottomrule
  \end{tabular}
  \vspace{-15pt}
\end{table}

\subsection{Robustness under Purification Attacks}
Another important category of post-processing operations that can impact robustness of protection includes regeneration or purification-style methods (e.g., DiffPure~\cite{nie2022DiffPure} and IMPRESS~\cite{cao2023impress}). These are different from common transformation-based and semantic-based edits which are typically applied without assuming knowledge of the protection. Purification is performed by informed adversaries who explicitly attempt to remove the protective signal before further use. Although such stress tests are informative, they answer a different question: whether protection survives targeted removal. PREEMPT instead asks how long protection remains effective under a realistic distribution and editing lifecycle after release. 
Results evaluating resistance to targeted purification (DiffPure and IMPRESS) are included in SM for completeness.

\section{Conclusion}
\label{sec:conclusion}
In this paper, we present PREEMPT, a holistic evaluation framework for assessing the robustness of proactive image protection methods under realistic editing scenarios. 
Rather than attempting to enumerate all possible editing chains, PREEMPT evaluates robustness through representative one-shot and sequential editing scenarios that provide a practical and reproducible evaluation protocol.
We introduce unified trajectory-based metrics
to characterize robustness evolution across different protection paradigms.
Beyond evaluating robustness, PREEMPT reveals three key insights. First, robustness evolves differently across protection paradigms. 
Second, similar one-shot robustness does not necessarily imply comparable long-horizon robustness, as methods may follow different degradation trajectories under sequential editing. 
Third, robustness degradation is largely concentrated in the initial semantic edit, highlighting its dominant role in long-horizon robustness evolution. These insights suggest that protection methods should optimize robustness over editing trajectories rather than a single operating point. 
We hope that PREEMPT provides a foundation for future proactive image protection \mbox{research}.
{
    \small
    \bibliographystyle{ieeenat_fullname}
    \bibliography{main}

@article{liu2023generating,
  title={Generating robust adversarial examples against online social networks (osns)},
  author={Liu, Jun and Zhou, Jiantao and Wu, Haiwei and Sun, Weiwei and Tian, Jinyu},
  journal={ACM Transactions on Multimedia Computing, Communications and Applications},
  volume={20},
  number={4},
  pages={1--26},
  year={2023},
  publisher={ACM New York, NY}
}

@String(CVPR= {IEEE Conf. Comput. Vis. Pattern Recog.})

@String(ICCV= {Int. Conf. Comput. Vis.})

@String(ECCV= {Eur. Conf. Comput. Vis.})

@String(ICIP = {IEEE Int. Conf. Image Process.})

@String(ICLR = {Int. Conf. Learn. Represent.})

@String(AAAI = {AAAI})

@String(CVPR  = {CVPR})

@String(ICCV  = {ICCV})

@String(ECCV  = {ECCV})

@String(ICIP  = {ICIP})

@String(ICLR  = {ICLR})

@inproceedings{ruiz2023dreambooth,
  title={Dreambooth: Fine tuning text-to-image diffusion models for subject-driven generation},
  author={Ruiz, Nataniel and Li, Yuanzhen and Jampani, Varun and Pritch, Yael and Rubinstein, Michael and Aberman, Kfir},
  booktitle={Proceedings of the IEEE/CVF Conference on Computer Vision and Pattern Recognition (CVPR)},
  pages={22500--22510},
  year={2023}
}

@article{ho2020denoising,
  title={Denoising diffusion probabilistic models},
  author={Ho, Jonathan and Jain, Ajay and Abbeel, Pieter},
  journal={Advances in neural information processing systems},
  volume={33},
  pages={6840--6851},
  year={2020}
}

@inproceedings{
gal2022image,
title={An Image is Worth One Word: Personalizing Text-to-Image Generation using Textual Inversion},
author={Rinon Gal and Yuval Alaluf and Yuval Atzmon and Or Patashnik and Amit Haim Bermano and Gal Chechik and Daniel Cohen-or},
booktitle={The Eleventh International Conference on Learning Representations (ICLR)},
year={2023}
}

@inproceedings{xue2023toward,
  title={Toward effective protection against diffusion-based mimicry through score distillation},
  author={Xue, Haotian and Liang, Chumeng and Wu, Xiaoyu and Chen, Yongxin},
  booktitle={International Conference on Learning Representations (ICLR)},
  year={2024}
}

@inproceedings{xu2024perturbing,
  title={Perturbing Attention Gives You More Bang for the Buck: Subtle Imaging Perturbations That Efficiently Fool Customized Diffusion Models},
  author={Xu, Jingyao and Lu, Yuetong and Li, Yandong and Lu, Siyang and Wang, Dongdong and Wei, Xiang},
  booktitle={Proceedings of the IEEE/CVF Conference on Computer Vision and Pattern Recognition (CVPR)},
  pages={24534--24543},
  year={2024}
}

@inproceedings{hessel2021clipscore,
    title = "{CLIPS}core: A Reference-free Evaluation Metric for Image Captioning",
    author = "Hessel, Jack  and
      Holtzman, Ari  and
      Forbes, Maxwell  and
      Le Bras, Ronan  and
      Choi, Yejin",
    booktitle = "Conference on Empirical Methods in Natural Language Processing",
    year = "2021",
    publisher = "Association for Computational Linguistics",
    pages = "7514--7528",
}

@misc{lei2024diffusetrace,
  title={Diffusetrace: A transparent and flexible watermarking scheme for latent diffusion model},
  author={Lei, Liangqi and Gai, Keke and Yu, Jing and Zhu, Liehuang},
  eprint="2405.02696",
  archivePrefix="arXiv",
  year={2024}
}

@inproceedings{
lu2024robust,
title={Robust Watermarking Using Generative Priors Against Image Editing: From Benchmarking to Advances},
author={Shilin Lu and Zihan Zhou and Jiayou Lu and Yuanzhi Zhu and Adams Wai-Kin Kong},
booktitle={The Thirteenth International Conference on Learning Representations},
year={2025}
}

@inproceedings{rombach2022high,
  title={High-resolution image synthesis with latent diffusion models. 2022 IEEE},
  author={Rombach, Robin and Blattmann, Andreas and Lorenz, Dominik and Esser, Patrick and Ommer, Bj{\"o}rn},
  booktitle={CVF Conference on Computer Vision and Pattern Recognition (CVPR)},
  volume={1},
  year={2021}
}

@article{artgan2018,
  title={Improved ArtGAN for Conditional Synthesis of Natural Image and Artwork},
  author={Tan, Wei Ren and Chan, Chee Seng and Aguirre, Hernan and Tanaka, Kiyoshi},
  journal={IEEE Transactions on Image Processing},
  volume    = {28},
  number    = {1},
  pages     = {394--409},
  year      = {2019},
  url       = {https://doi.org/10.1109/TIP.2018.2866698},
  doi       = {10.1109/TIP.2018.2866698}
}

@inproceedings{
ye2024duaw,
title={{DUAW}: Data-free Universal Adversarial Watermark against Stable Diffusion Customization},
author={Xiaoyu Ye and Hao Huang and Jiaqi An and Yongtao Wang},
booktitle={ICLR 2024 Workshop on Secure and Trustworthy Large Language Models},
year={2024},
url={https://openreview.net/forum?id=XYD342nKy8}
}

@inproceedings{
wen2023treerings,
title={Tree-Rings Watermarks: Invisible Fingerprints for Diffusion Images},
author={Yuxin Wen and John Kirchenbauer and Jonas Geiping and Tom Goldstein},
booktitle={Thirty-seventh Conference on Neural Information Processing Systems},
year={2023},
url={https://openreview.net/forum?id=Z57JrmubNl}
}

@article{heusel2017gans,
  title={Gans trained by a two time-scale update rule converge to a local nash equilibrium},
  author={Heusel, Martin and Ramsauer, Hubert and Unterthiner, Thomas and Nessler, Bernhard and Hochreiter, Sepp},
  journal={Advances in Neural Information Processing Systems},
  volume={30},
  year={2017}
}

@InProceedings{wang2024edit,
    author = {Wang, Hanhui and Zhang, Yihua and Bai, Ruizheng and Zhao, Yue and Liu, Sijia and Tu, Zhengzhong},
    title = {Edit Away and My Face Will not Stay: Personal Biometric Defense against Malicious Generative Editing},
    booktitle = {Proceedings of the IEEE/CVF Conference on Computer Vision and Pattern Recognition (CVPR)},
    month     = {June},
    year      = {2025}
}

@article{somepalli2024measuring,
title={Measuring Style Similarity in Diffusion Models},
author={Somepalli, Gowthami and Gupta, Anubhav and Gupta, Kamal and Palta, Shramay and Goldblum, Micah and Geiping, Jonas and Shrivastava, Abhinav and Goldstein, Tom},
journal={Proceedings of the European conference on computer vision (ECCV)},
year={2024}
}

@article{tang2025styleprotect,
  title={StyleProtect: Safeguarding Artistic Identity in Fine-tuned Diffusion Models},
  author={Tang, Qiuyu and Krinsky, Joshua and Bharati, Aparna},
  journal={arXiv preprint arXiv:2509.13711},
  year={2025}
}

@InProceedings{salman2023raising,
  title = 	 {Raising the Cost of Malicious {AI}-Powered Image Editing},
  author =       {Salman, Hadi and Khaddaj, Alaa and Leclerc, Guillaume and Ilyas, Andrew and Madry, Aleksander},
  booktitle = 	 {Proceedings of the 40th International Conference on Machine Learning},
  pages = 	 {29894--29918},
  year = 	 {2023},
  editor = 	 {Krause, Andreas and Brunskill, Emma and Cho, Kyunghyun and Engelhardt, Barbara and Sabato, Sivan and Scarlett, Jonathan},
  volume = 	 {202},
  series = 	 {Proceedings of Machine Learning Research},
  month = 	 {23--29 Jul},
  publisher =    {PMLR},
  url = 	 {https://proceedings.mlr.press/v202/salman23a.html}
}

@inproceedings{shan2023glaze,
  title={Glaze: Protecting artists from style mimicry by Text-to-Image models},
  author={Shan, Shawn and Cryan, Jenna and Wenger, Emily and Zheng, Haitao and Hanocka, Rana and Zhao, Ben Y},
  booktitle={32nd USENIX Security Symposium (USENIX Security 23)},
  pages={2187--2204},
  year={2023}
}

@inproceedings{liu2024metacloak,
  title={MetaCloak: Preventing Unauthorized Subject-driven Text-to-image Diffusion-based Synthesis via Meta-learning},
  author={Liu, Yixin and Fan, Chenrui and Dai, Yutong and Chen, Xun and Zhou, Pan and Sun, Lichao},
  booktitle={IEEE/CVF Conference on Computer Vision and Pattern Recognition},
  pages={24219--24228},
  year={2024}
}

@article{liang2023mist,
  title={Mist: Towards improved adversarial examples for diffusion models},
  author={Liang, Chumeng and Wu, Xiaoyu},
  journal={arXiv preprint arXiv:2305.12683},
  year={2023}
}

@inproceedings{zhang2024editguard,
  title={Editguard: Versatile image watermarking for tamper localization and copyright protection},
  author={Zhang, Xuanyu and Li, Runyi and Yu, Jiwen and Xu, Youmin and Li, Weiqi and Zhang, Jian},
  booktitle={Proceedings of the IEEE/CVF Conference on Computer Vision and Pattern Recognition (CVPR)},
  pages={11964--11974},
  year={2024}
}

@inproceedings{
hu2022lora,
title={Lo{RA}: Low-Rank Adaptation of Large Language Models},
author={Edward J Hu and Yelong Shen and Phillip Wallis and Zeyuan Allen-Zhu and Yuanzhi Li and Shean Wang and Lu Wang and Weizhu Chen},
booktitle={International Conference on Learning Representations (ICLR)},
year={2022},
url={https://openreview.net/forum?id=nZeVKeeFYf9}
}

@inproceedings{fernandez2023stable,
  title={The stable signature: Rooting watermarks in latent diffusion models},
  author={Fernandez, Pierre and Couairon, Guillaume and J{\'e}gou, Herv{\'e} and Douze, Matthijs and Furon, Teddy},
  booktitle={Proceedings of the IEEE/CVF International Conference on Computer Vision (CVPR)},
  pages={22466--22477},
  year={2023}
}

@inproceedings{
song2021denoising,
title={Denoising Diffusion Implicit Models},
author={Jiaming Song and Chenlin Meng and Stefano Ermon},
booktitle={International Conference on Learning Representations (ICLR)},
year={2021},
url={https://openreview.net/forum?id=St1giarCHLP}
}

@inproceedings{van2023anti,
  title={Anti-dreambooth: Protecting users from personalized text-to-image synthesis},
  author={Van Le, Thanh and Phung, Hao and Nguyen, Thuan Hoang and Dao, Quan and Tran, Ngoc N and Tran, Anh},
  booktitle={IEEE/CVF International Conference on Computer Vision (ICCV)},
  pages={2116--2127},
  year={2023}
}

@article{cui2025diffusionshield,
author = {Cui, Yingqian and Ren, Jie and Xu, Han and He, Pengfei and Liu, Hui and Sun, Lichao and Xing, Yue and Tang, Jiliang},
title = {DiffusionShield: A Watermark for Data Copyright Protection against Generative Diffusion Models},
year = {2025},
publisher = {Association for Computing Machinery},
volume = {26},
number = {2},
issn = {1931-0145},
journal = {SIGKDD Explor. Newsl.},
pages = {60–75},
numpages = {16}
}

@inproceedings{zhang2018unreasonable,
  title={The unreasonable effectiveness of deep features as a perceptual metric},
  author={Zhang, Richard and Isola, Phillip and Efros, Alexei A and Shechtman, Eli and Wang, Oliver},
  booktitle={IEEE Conference on Computer Vision and Pattern Recognition (CVPR)},
  pages={586--595},
  year={2018}
}

@inproceedings{hu2025robust,
  title={Robust-wide: Robust watermarking against instruction-driven image editing},
  author={Hu, Runyi and Zhang, Jie and Xu, Ting and Li, Jiwei and Zhang, Tianwei},
  booktitle={European Conference on Computer Vision},
  pages={20--37},
  year={2025},
  organization={Springer}
}

@inproceedings{Trustmark-ICCV-2025,
  title = {TrustMark: Robust Watermarking and Watermark Removal for Arbitrary Resolution Images},
  author={Bui, Tu and Agarwal, Shruti and Collomosse, John},
  booktitle = {IEEE International Conference on Computer Vision (ICCV)},
  year = {2025},
  month = oct
}

@article{li2025editthinker,
  title={Editthinker: Unlocking iterative reasoning for any image editor},
  author={Li, Hongyu and Zhang, Manyuan and Zheng, Dian and Guo, Ziyu and Jia, Yimeng and Feng, Kaituo and Yu, Hao and Liu, Yexin and Feng, Yan and Pei, Peng and others},
  journal={arXiv preprint arXiv:2512.05965},
  year={2025}
}

@article{chen2025edival,
  title={Edival-agent: An object-centric framework for automated, fine-grained evaluation of multi-turn editing},
  author={Chen, Tianyu and Zhang, Yasi and Zhang, Zhi and Yu, Peiyu and Wang, Shu and Wang, Zhendong and Lin, Kevin and Wang, Xiaofei and Yang, Zhengyuan and Li, Linjie and others},
  journal={arXiv preprint arXiv:2509.13399},
  year={2025}
}

@inproceedings{10.1145/2856400.2856417,
author = {Chen, Hsiang-Ting and Wei, Li-Yi and Hartmann, Bj{\"o}rn and Agrawala, Maneesh},
title = {Data-driven adaptive history for image editing},
year = {2016},
isbn = {9781450340434},
publisher = {Association for Computing Machinery},
address = {New York, NY, USA},
url = {https://doi.org/10.1145/2856400.2856417},
doi = {10.1145/2856400.2856417},
booktitle = {Proceedings of the 20th ACM SIGGRAPH Symposium on Interactive 3D Graphics and Games},
pages = {103–111},
numpages = {9},
location = {Redmond, Washington},
series = {I3D '16}
}

@inproceedings{cao2023impress,
  title     = {{IMPRESS}: Evaluating the Resilience of Imperceptible
               Perturbations Against Unauthorized Data Usage in
               Diffusion-Based Generative {AI}},
  author    = {Bochuan Cao and Changjiang Li and Ting Wang and
               Jinyuan Jia and Bo Li and Jinghui Chen},
  booktitle = {Advances in Neural Information Processing Systems},
  year      = {2023},
  url       = {https://arxiv.org/abs/2310.19248}
}

@inproceedings{liang2023adversarial,
  title={Adversarial example does good: Preventing painting imitation from diffusion models via adversarial examples},
  author={Liang, Chumeng and Wu, Xiaoyu and Hua, Yang and Zhang, Jiaru and Xue, Yiming and Song, Tao and Xue, Zhengui and Ma, Ruhui and Guan, Haibing},
  booktitle={International Conference on Machine Learning (ICML)},
  pages={20763--20786},
  year={2023},
  organization={PMLR}
}

@inproceedings{deng2019arcface,
  title={Arcface: Additive angular margin loss for deep face recognition},
  author={Deng, Jiankang and Guo, Jia and Xue, Niannan and Zafeiriou, Stefanos},
  booktitle={Proceedings of the IEEE/CVF Conference on Computer Vision and Pattern Recognition (CVPR)},
  pages={4690--4699},
  year={2019}
}

@misc{openai2023gpt4,
  author       = {OpenAI},
  title        = {GPT-4 Technical Report},
  year         = {2023},
  eprint="2303.08774",
  archivePrefix="arXiv",
}

@inproceedings{
karras2018progressive,
title={Progressive Growing of {GAN}s for Improved Quality, Stability, and Variation},
author={Tero Karras and Timo Aila and Samuli Laine and Jaakko Lehtinen},
booktitle={International Conference on Learning Representations (ICLR)},
year={2018},
url={https://openreview.net/forum?id=Hk99zCeAb},
}

@inproceedings{brooks2023instructpix2pix,
  title={Instructpix2pix: Learning to follow image editing instructions},
  author={Brooks, Tim and Holynski, Aleksander and Efros, Alexei A},
  booktitle={Proceedings of the IEEE/CVF conference on computer vision and pattern recognition (CVPR)},
  pages={18392--18402},
  year={2023}
}

@article{tang2025perturbation,
  title={Is Perturbation-Based Image Protection Disruptive to Image Editing?},
  author={Tang, Qiuyu and Ayambem, Bonor and Chuah, Mooi Choo and Bharati, Aparna},
  journal={2025 IEEE International Conference on Image Processing (ICIP)},
  year={2025}
}

@inproceedings{xu2025lmm4edit,
  title={Lmm4edit: Benchmarking and evaluating multimodal image editing with lmms},
  author={Xu, Zitong and Duan, Huiyu and Liu, Bingnan and Ma, Guangji and Wang, Jiarui and Yang, Liu and Gao, Shiqi and Wang, Xiaoyu and Wang, Jia and Min, Xiongkuo and others},
  booktitle={Proceedings of the 33rd ACM International Conference on Multimedia},
  pages={6908--6917},
  year={2025}
}

@article{verdoliva2020media,
  title={Media forensics and deepfakes: an overview},
  author={Verdoliva, Luisa},
  journal={IEEE journal of selected topics in signal processing},
  volume={14},
  number={5},
  pages={910--932},
  year={2020},
  publisher={IEEE}
}

@inproceedings{sander2025watermark,
  title={Watermark Anything with Localized Messages},
  author={Sander, Tom and Fernandez, Pierre and Durmus, Alain and Furon, Teddy and Douze, Matthijs},
  booktitle={International Conference on Learning Representations (ICLR)},
  year={2025}
}

@inproceedings{
rabbani2024waves,
title={{WAVES}: Benchmarking the Robustness of Image Watermarks},
author={Tahseen Rabbani and Bang An and Mucong Ding and Aakriti Agrawal and Yuancheng Xu and Chenghao Deng and Sicheng Zhu and Abdirisak Mohamed and Yuxin Wen and Tom Goldstein and Furong Huang},
booktitle={Privacy Regulation and Protection in Machine Learning},
year={2024},
url={https://openreview.net/forum?id=tUumjNa7TF}
}

@InProceedings{Tancik_2020_CVPR,
author = {Tancik, Matthew and Mildenhall, Ben and Ng, Ren},
title = {StegaStamp: Invisible Hyperlinks in Physical Photographs},
booktitle = {Proceedings of the IEEE/CVF Conference on Computer Vision and Pattern Recognition (CVPR)},
month = {June},
year = {2020}
}

@article{oquab2023dinov2,
  title={Dinov2: Learning robust visual features without supervision},
  author={Oquab, Maxime and Darcet, Timoth{\'e}e and Moutakanni, Th{\'e}o and Vo, Huy and Szafraniec, Marc and Khalidov, Vasil and Fernandez, Pierre and Haziza, Daniel and Massa, Francisco and El-Nouby, Alaaeldin and others},
  journal={arXiv preprint arXiv:2304.07193},
  year={2023}
}

@inproceedings{
tang2025watermarks,
title={Watermarks vs. Perturbations for Preventing {AI}-based Style Editing},
author={Qiuyu Tang and Aparna Bharati},
booktitle={The 1st Workshop on GenAI Watermarking},
year={2025},
url={https://openreview.net/forum?id=mRCXybDMF6}
}

@article{couairon2022diffedit,
  title={Diffedit: Diffusion-based semantic image editing with mask guidance},
  author={Couairon, Guillaume and Verbeek, Jakob and Schwenk, Holger and Cord, Matthieu},
  journal={arXiv preprint arXiv:2210.11427},
  year={2022}
}

@inproceedings{liu2025watermarking,
  title={Watermarking One for All: A Robust Watermarking Scheme Against Partial Image Theft},
  author={Liu, Gaozhi and Cao, Silu and Qian, Zhenxing and Zhang, Xinpeng and Li, Sheng and Peng, Wanli},
  booktitle={Proceedings of the Computer Vision and Pattern Recognition Conference},
  pages={8225--8234},
  year={2025}
}

@inproceedings{jeong2025faceshield,
  title={Faceshield: Defending facial image against deepfake threats},
  author={Jeong, Jaehwan and In, Sumin and Kim, Sieun and Shin, Hannie and Jeong, Jongheon and Yoon, Sang Ho and Chung, Jaewook and Kim, Sangpil},
  booktitle={Proceedings of the IEEE/CVF International Conference on Computer Vision},
  pages={10364--10374},
  year={2025}
}

@inproceedings{mi2024visual,
    author = {Mi, Xiaoyue and Tang, Fan and Cao, Juan and Li, Peng and Liu, Yang},
    title = {Visual-friendly concept protection via selective adversarial perturbations},
    booktitle ={Proceedings of the AAAI Conference on Artificial Intelligence},
    year = {2024},
    volume    = {40},
    number    = {46},
    pages     = {39033--39041},
    doi       = {10.1609/aaai.v40i46.41250}
}

@inproceedings{liu2015deep,
  title={Deep learning face attributes in the wild},
  author={Liu, Ziwei and Luo, Ping and Wang, Xiaogang and Tang, Xiaoou},
  booktitle={Proceedings of the IEEE international conference on computer vision},
  pages={3730--3738},
  year={2015}
}

@inproceedings{nie2022DiffPure,
  title={Diffusion Models for Adversarial Purification},
  author={Nie, Weili and Guo, Brandon and Huang, Yujia and Xiao, Chaowei and Vahdat, Arash and Anandkumar, Anima},
  booktitle = {International Conference on Machine Learning (ICML)},
  year={2022}
}
}

\end{document}


\maketitle

\section{Experimental Details}

\subsection{Dataset}
PREEMPT is evaluated on a multi-domain benchmark constructed from WikiArt~\cite{artgan2018}, CelebA-HQ~\cite{liu2015deep,karras2018progressive}, and DreamBooth~\cite{ruiz2023dreambooth}, covering artistic, biometric, and object-centric domains. All images are resized to $512\times512$ before protection and editing. Dataset statistics are summarized in Tab.~\ref{tab:dataset_stats}.

\begin{table}[htb]
\centering
\caption{Dataset statistics. }
\label{tab:dataset_stats}
\begin{tabular}{lcccl}
\toprule
Subdataset & Domain & $N$ & Pool \\
\midrule
Faces  & Identity & 30 & CelebA-HQ   \\
Arts   & Style    & 30 & WikiArt  \\
Objects& Object   & 40 & DreamBooth   \\
\midrule
\textbf{Total} & --- & \textbf{100} & --- \\
\bottomrule
\end{tabular}
\end{table}

\paragraph{Selection criteria.}
Images are selected according to the following principles:
\begin{enumerate}
    \item \textbf{Domain consistency.} Each subdataset represents a single concept type (identity, style, or object), ensuring that the corresponding evaluation metric (ArcFace, CSD, or DINO) matches the protected content.
    \item \textbf{Prompt completeness.} Every image supports the full set of semantic editing prompts for all evaluation levels.
    \item \textbf{Protection completeness.} Only images available under all protection methods are retained.
    \item \textbf{Uniform preprocessing.} All images are converted to RGB and resized to $512\times512$ before protection and editing.
\end{enumerate}

Each image is protected by seven methods (Clean, AdvDM, Mist, PhotoGuard, SDST, Robust-Wide, TrustMark, VINE, and WAM), yielding 700 protected images for each evaluation stage.

\subsection{Editing Configuration}
\paragraph{Transformation-based edit.}
Transformation-based edits consist of Gaussian blur ($\sigma=2.0$), JPEG compression (quality factors 50 and 70), crop-and-resize (cropping 85\% of the image area followed by resizing to the original resolution), horizontal flipping, and brightness/contrast adjustment with a scaling factor of 1.5.

\begin{tcolorbox}[title=Prompt Template for Semantic Edit Generation]

\textbf{Input:}
Image $I$

\textbf{Task:}
Analyze the image and generate realistic editing prompts.

\textbf{Stage 1. Visual Understanding}
\begin{itemize}
    \item Main subjects
    \item Attributes
    \item Artistic style
    \item Spatial layout
\end{itemize}

\textbf{Stage 2. Domain Awareness}
\begin{itemize}
    \item Face $\rightarrow$ identity manipulation
    \item Artwork $\rightarrow$ object/scene manipulation
    \item Natural image $\rightarrow$ object manipulation
\end{itemize}

\textbf{Stage 3. Prompt Generation}

Categories:
\[
\mathcal{C}=\{\texttt{Content},\texttt{Attribute},\texttt{Style},\texttt{Spatial}\}.
\]

Constraints:
\begin{enumerate}
    \item At most four prompts per category.
    \item Grounded edits only.
    \item Short and imperative phrases.
    \item Ensure diversity and non-redundancy.
\end{enumerate}

\textbf{Output}
\begin{verbatim}
{
  "Content": [...],
  "Attribute": [...],
  "Style": [...],
  "Spatial": [...]
}
\end{verbatim}

\end{tcolorbox}

\paragraph{Semantic edit.}
We explain how to generate prompts for multiple semantic edits. Semantic edits are organized into four categories,
$C=\{\text{Content}, \text{Attribute}, \text{Style}, \text{Spatial}\}$.
For each input image, GPT-4.1-mini~\cite{openai2023gpt4} generates four category-specific editing prompts. Prompt generation follows three principles: (1) edits should be visually grounded in the original image; (2) prompts should cover diverse semantic modifications; and (3) redundant instructions are avoided. Example prompts for each category are provided in Tab.~\ref{tab:qualitative_editing_prompts}.

\begin{table*}[h]
\centering
\caption{Examples of editing prompts for various categories.}
\label{tab:qualitative_editing_prompts}
\begin{tabular}{llcl}
\toprule
Category & Type & Editing prompt \\
\midrule
Face     &  Content         & Change the person's hair color to brunette \\
Face     &  Style           & Render the portrait in a watercolor painting style \\
Face     &  Spatial         & Zoom in closer on the face \\
\addlinespace
Art      & Spatial         & Zoom in closer on the water lilies \\
Art      & Style           & Render in a more impressionist style with visible brush strokes \\
\addlinespace
Object   & Style           & Render the image as a watercolor painting \\
Object   & Attribute       & Change lens color to blue tint \\
\bottomrule
\end{tabular}
\end{table*}

Generated prompts are executed using Stable Diffusion v2.1~\cite{rombach2022high} and InstructPix2Pix~\cite{brooks2023instructpix2pix} under their default inference settings.

\subsection{Randomization}
Unless otherwise specified, all random components use a fixed seed of 42. For sequential editing, edit operations are sampled using the same seed to ensure reproducibility.

\section{Details of One-shot Editing}

\subsection{One-shot transformation-based Edit}

To illustrate domain-specific robustness, we report concept similarity under image-transformation-based edits separately for the face, artwork, and object subsets in Tab.~\ref{tab:l1_concept_similarity_domains}.

\begin{table*}[t]
\centering

\begin{subtable}[t]{\linewidth}
\centering
\caption{Face subset}
\label{tab:l1_concept_similarity_faces}
\begin{tabular}{lcccccc|c}
\toprule
Protection & JPEG (Q=70) & JPEG (Q=50) & Bright./Contrast & Gaussian Blur & Crop \& Resize & Horiz.\ Flip & Avg. \\
\midrule
\textit{None}  & 0.989 & 0.977 & 0.972 & 0.980 & 0.908 & 0.933 & 0.960 \\
AdvDM          & 0.942 & 0.935 & 0.927 & 0.954 & 0.891 & 0.896 & 0.924 \\
MIST           & 0.905 & 0.897 & 0.887 & 0.906 & 0.842 & 0.868 & 0.884 \\
PhotoGuard     & 0.904 & 0.898 & 0.889 & 0.903 & 0.841 & 0.870 & 0.884 \\
SDST           & 0.913 & 0.910 & 0.896 & 0.913 & 0.851 & 0.872 & 0.892 \\
\bottomrule
\end{tabular}
\end{subtable}

\vspace{0.8em}

\begin{subtable}[t]{\linewidth}
\centering
\caption{Artwork subset}
\label{tab:l1_concept_similarity_arts}
\begin{tabular}{lcccccc|c}
\toprule
Protection & JPEG (Q=70) & JPEG (Q=50) & Bright./Contrast & Gaussian Blur & Crop \& Resize & Horiz.\ Flip & Avg. \\
\midrule
\textit{None}  & 0.975 & 0.945 & 0.957 & 0.927 & 0.940 & 0.978 & 0.954 \\
AdvDM          & 0.734 & 0.740 & 0.729 & 0.758 & 0.657 & 0.734 & 0.726 \\
MIST           & 0.835 & 0.830 & 0.829 & 0.815 & 0.802 & 0.842 & 0.826 \\
PhotoGuard     & 0.838 & 0.832 & 0.828 & 0.813 & 0.806 & 0.843 & 0.827 \\
SDST           & 0.840 & 0.838 & 0.842 & 0.809 & 0.826 & 0.856 & 0.835 \\
\bottomrule
\end{tabular}
\end{subtable}

\vspace{0.8em}

\begin{subtable}[t]{\linewidth}
\centering
\caption{Object subset}
\label{tab:l1_concept_similarity_objects}
\begin{tabular}{lcccccc|c}
\toprule
Protection & JPEG (Q=70) & JPEG (Q=50) & Bright./Contrast & Gaussian Blur & Crop \& Resize & Horiz.\ Flip & Avg. \\
\midrule
\textit{None}  & 0.998 & 0.996 & 0.992 & 0.997 & 0.964 & 0.983 & 0.988 \\
AdvDM          & 0.927 & 0.924 & 0.930 & 0.917 & 0.898 & 0.922 & 0.920 \\
MIST           & 0.932 & 0.927 & 0.937 & 0.913 & 0.905 & 0.926 & 0.923 \\
PhotoGuard     & 0.931 & 0.929 & 0.937 & 0.912 & 0.904 & 0.927 & 0.923 \\
SDST           & 0.932 & 0.929 & 0.936 & 0.918 & 0.904 & 0.929 & 0.925 \\
\bottomrule
\end{tabular}
\end{subtable}

\caption{Semantic similarity under image-transformation-based edits for the three evaluation domains. Lower values indicate stronger protection robustness of all perturbation-based methods after image transformations.}
\label{tab:l1_concept_similarity_domains}
\end{table*}

\subsection{One-shot Semantic Edit.}

We report one-shot semantic editing results using Stable Diffusion 2.1~\cite{rombach2022high} here, complementing the InstructPix2Pix~\cite{brooks2023instructpix2pix} results presented in the main paper. Table~\ref{tab:l2_semantic_similarity_sd21} summarizes the semantic similarity after one-shot edits. Compared with unprotected images, all proactive protection methods substantially reduce semantic similarity, demonstrating that they remain effective under semantic manipulation. Among the evaluated methods, SDST~\cite{xue2023toward} achieves the strongest protection performance, followed closely by MIST~\cite{liang2023mist} and PhotoGuard~\cite{salman2023raising}, while AdvDM~\cite{liang2023adversarial} is comparatively less effective.
The overall conclusion is consistent across editing models. This result reinforces the importance of evaluating protection methods under diverse editing models rather than relying on a single editing pipeline.

\begin{table*}[t]
\centering
\caption{Semantic similarity under one-shot semantic edits using Stable Diffusion 2.1.}
\label{tab:l2_semantic_similarity_sd21}
\begin{tabular}{lcccc|c}
\toprule
Protection & Content & Attribute & Style & Spatial & Average \\
\midrule
\textit{None}        & 0.298 & 0.261 & 0.170 & 0.268 & 0.249 \\
AdvDM         & 0.169 & 0.118 & 0.077 & 0.139 & 0.126 \\
MIST          & 0.130 & 0.086 & 0.051 & 0.101 & 0.092 \\
PhotoGuard    & 0.127 & 0.093 & 0.049 & 0.103 & 0.093 \\
SDST         & 0.123 & 0.089 & 0.051 & 0.101 & 0.091 \\
\bottomrule
\end{tabular}
\end{table*}

\section{Details of Sequential Editing}

\subsection{Clean-Trajectory Control under Semantic Edits}
As defined in the main paper, for perturbation-based methods we apply the same semantic/sequential editing chains to unprotected images and report
$\Delta(d)=S_{\mathrm{clean}}(d)-S_{\mathrm{prot}}(d)$.
Watermarking methods are excluded, since their evaluation uses bit accuracy rather than semantic similarity.

\begin{figure}[htb]
    \centering
    \includegraphics[width=\linewidth]{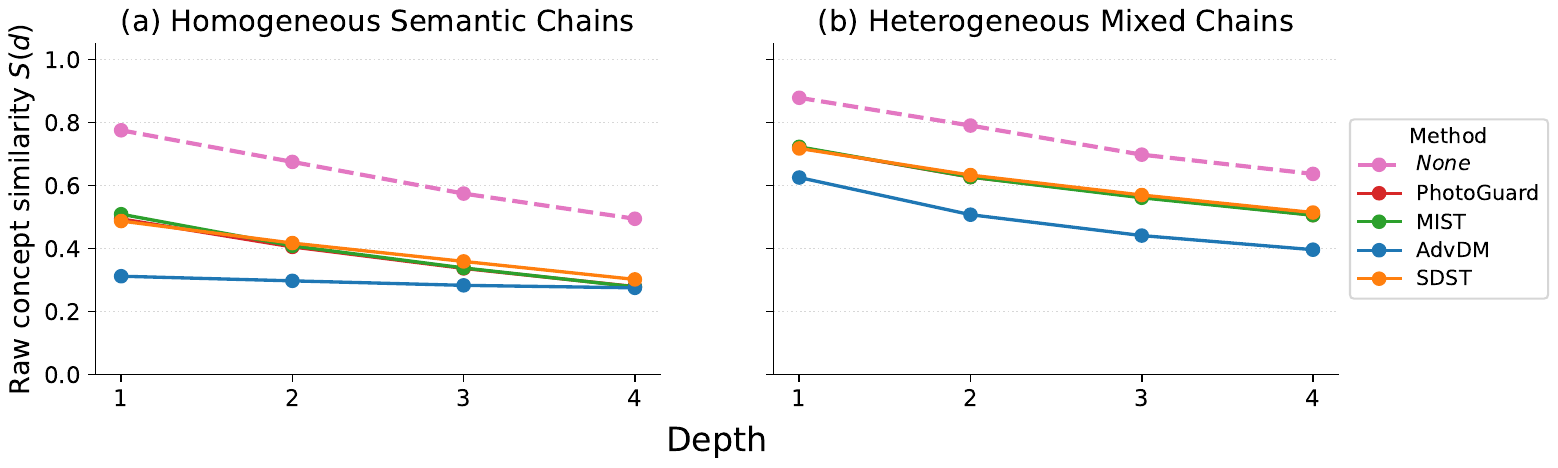}
    \caption{Clean-trajectory control under InstructPix2Pix editing: raw concept similarity $S(d)$ for unprotected (``None'') vs.\ perturbation-protected images on the same chains. The clean baseline remains high at depth~4 ($S(4)\approx0.49$/$0.64$), while protections yield $\Delta=S_{\mathrm{clean}}-S_{\mathrm{prot}}>0$.}
    \label{fig:clean}
\end{figure}

Fig.~\ref{fig:clean} plots raw concept similarity $S(d)$ for the clean trajectory (``None'') and for each perturbation method under InstructPix2Pix. Table~\ref{tab:supp-clean-delta} summarizes $\Delta$ at successive depths. Two observations follow. First, the clean baseline already declines with depth (e.g., $S(4)\approx0.49$ on homogeneous chains and $S(4)\approx0.64$ on heterogeneous chains), showing that semantic edits themselves induce substantial semantic drift. Second, all evaluated protections yield $\Delta(d)>0$ throughout the trajectory: relative to the matched clean edit path, protection pushes similarity down by an additional gap. Thus, low semantic scores under sequential semantic editing reflect both intended edit effects and an extra protection-induced reduction, which the clean-trajectory control separates.

\begin{table}[t]
\centering
\caption{Clean-trajectory control under InstructPix2Pix Editing.
$\Delta(d)=S_{\mathrm{clean}}(d)-S_{\mathrm{prot}}(d)$ for perturbation-based methods.
All entries are positive, indicating an additional protection-induced gap beyond edit-induced semantic drift.}
\label{tab:supp-clean-delta}
\setlength{\tabcolsep}{3pt}
\begin{tabular}{lcccc|cccc}
\toprule
& \multicolumn{4}{c|}{Homogeneous} & \multicolumn{4}{c}{Heterogeneous} \\
Method & $d{=}1$ & $d{=}2$ & $d{=}3$ & $d{=}4$ & $d{=}1$ & $d{=}2$ & $d{=}3$ & $d{=}4$ \\
\midrule
SDST       & 0.29 & 0.26 & 0.22 & 0.19 & 0.16 & 0.16 & 0.13 & 0.12 \\
MIST       & 0.27 & 0.27 & 0.24 & 0.22 & 0.16 & 0.16 & 0.14 & 0.13 \\
PhotoGuard & 0.28 & 0.27 & 0.24 & 0.22 & 0.16 & 0.16 & 0.13 & 0.13 \\
AdvDM      & 0.46 & 0.38 & 0.29 & 0.22 & 0.25 & 0.28 & 0.26 & 0.24 \\
\bottomrule
\end{tabular}
\end{table}

\subsection{Additional Results}

In the main paper, we present the protection survival curves under sequential semantic edits generated by InstructPix2Pix. Here, we additionally report the corresponding Protection AUC values in Tab.~\ref{tab:auc}, which summarize the overall robustness throughout the editing trajectory.

\begin{table}[h]
\centering
\caption{Protection survival AUC after sequential edits ($D=4$)}
\label{tab:auc}
\begin{tabular}{lccc}
\toprule
Method & Homo. & Hetero. & Avg. \\
\midrule
\textit{None} & 0.691 & 0.791 & 0.741 \\
SDST & 0.515 & 0.724 & 0.620 \\
MIST & 0.508 & 0.725 & 0.616 \\
PhotoGuard & 0.506 & 0.727 & 0.616 \\
AdvDM & 0.417 & 0.631 & 0.524 \\
\midrule
\midrule
Robust-Wide & 0.968 & 0.804 & 0.886 \\
TrustMark & 0.835 & 0.912 & 0.873 \\
VINE & 0.920 & 0.790 & 0.855 \\
WAM & 0.831 & 0.853 & 0.842 \\
\bottomrule
\end{tabular}
\end{table}

The interpretation of AUC depends on the protection paradigm. For perturbation-based methods, a lower AUC indicates stronger protection, as semantic similarity remains consistently low throughout the editing process. In contrast, for watermark-based methods, a higher AUC indicates better robustness, reflecting more reliable watermark retrieval under sequential edits.

\subsection{Sensitivity to chain length and edit ordering}

We further evaluate longer-horizon robustness by extending the editing depth to $d=10$. The protection survival curves continue to decline, following the same degradation trend observed at $d=4$. Interestingly, the PSR gap between watermark-based and perturbation-based methods becomes much smaller under longer editing chains, indicating similar long-horizon robustness across the two protection paradigms.

\begin{figure*}[t]
    \centering
    \includegraphics[width=\linewidth]{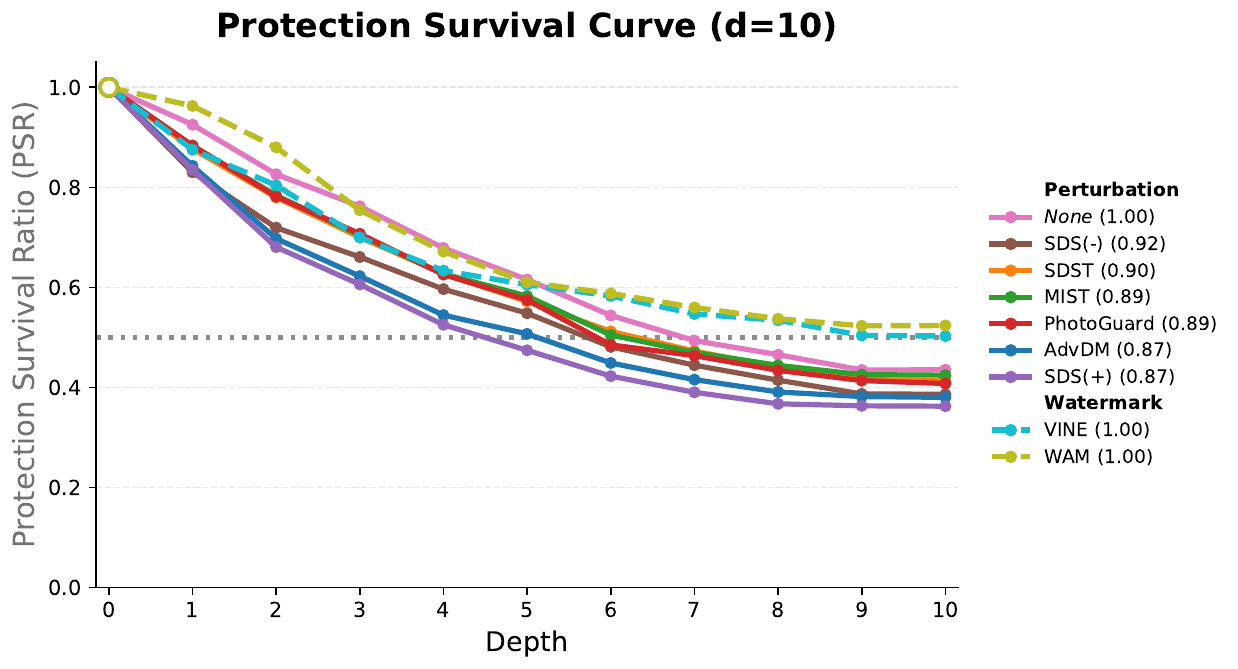}
    \caption{Protection Survival Curve under 10 steps' edits, covering all image-transformation operations and semantic edit types.}
    \label{fig:cx10}
\end{figure*}

Our experiments are conducted by randomly selecting from various editing types. We offer insights about different ordering of semantic-only and complex chains in the main paper Sec. 5.2. 

\subsection{One-Shot Robustness as a Predictor of Long-Horizon Survival}

We reports that one-shot robustness is only a weak predictor of long-horizon survival in the main paper. Here we detail the regression analysis and complementary visualizations.

\paragraph{Setup.}
For each perturbation-based protection method, the one-shot predictor is the mean concept similarity over the four semantic edit types (Content, Attribute, Style, Spatial) under InstructPix2Pix, matching the main protocol.
The long-horizon targets are computed on sequential chains of depth $D{=}4$:
(i)~normalized protection survival AUC,
$\mathrm{PAUC}=\mathrm{AUC}(\tilde{S})/D$;
and (ii)~$\tilde{S}(4)$ (PSR at depth~4).
We exclude the unprotected ``None'' control from the primary regression.
Unless noted otherwise, each point is one method$\times$chain pair (homogeneous and heterogeneous ; $n{=}48$).

\paragraph{Primary result.}
A linear regression from one-shot similarity to PAUC at the method$\times$chain level yields only a moderate association
($r{\approx}0.40$, $R^{2}{\approx}0.16$; Spearman $\rho{\approx}0.38$).
Thus one-shot scores explain little of the variance in sequential survival: methods with similar one-shot performance can follow substantially different multi-step trajectories.

\paragraph{Additional aggregations.}
Table~\ref{tab:oneshot_corr} compares correlation statistics under three aggregation units.
Image-level pooling ($n{=}600$) yields a stronger correlation, while method-level means ($n{=}6$) are statistically fragile due to the tiny sample size.
We therefore take \textbf{method$\times$chain} as the primary evidence for the claim in the main text.

\begin{table}[t]
\centering
\caption{Correlation between one-shot semantic similarity and long-horizon PAUC.
Primary claim uses method$\times$chain aggregation.}
\label{tab:oneshot_corr}
\begin{tabular}{lcccc}
\toprule
Unit & $n$ & Pearson $r$ & Spearman $\rho$ & $R^{2}$ \\
\midrule
Method$\times$chain & 48 & 0.397 & 0.382 & 0.158 \\
Method means & 6 & 0.995 & 0.886 & 0.991 \\
Image-level & 600 & 0.759 & 0.760 & 0.576 \\
\bottomrule
\end{tabular}
\end{table}

\begin{figure}[t]
\centering
\includegraphics[width=\linewidth]{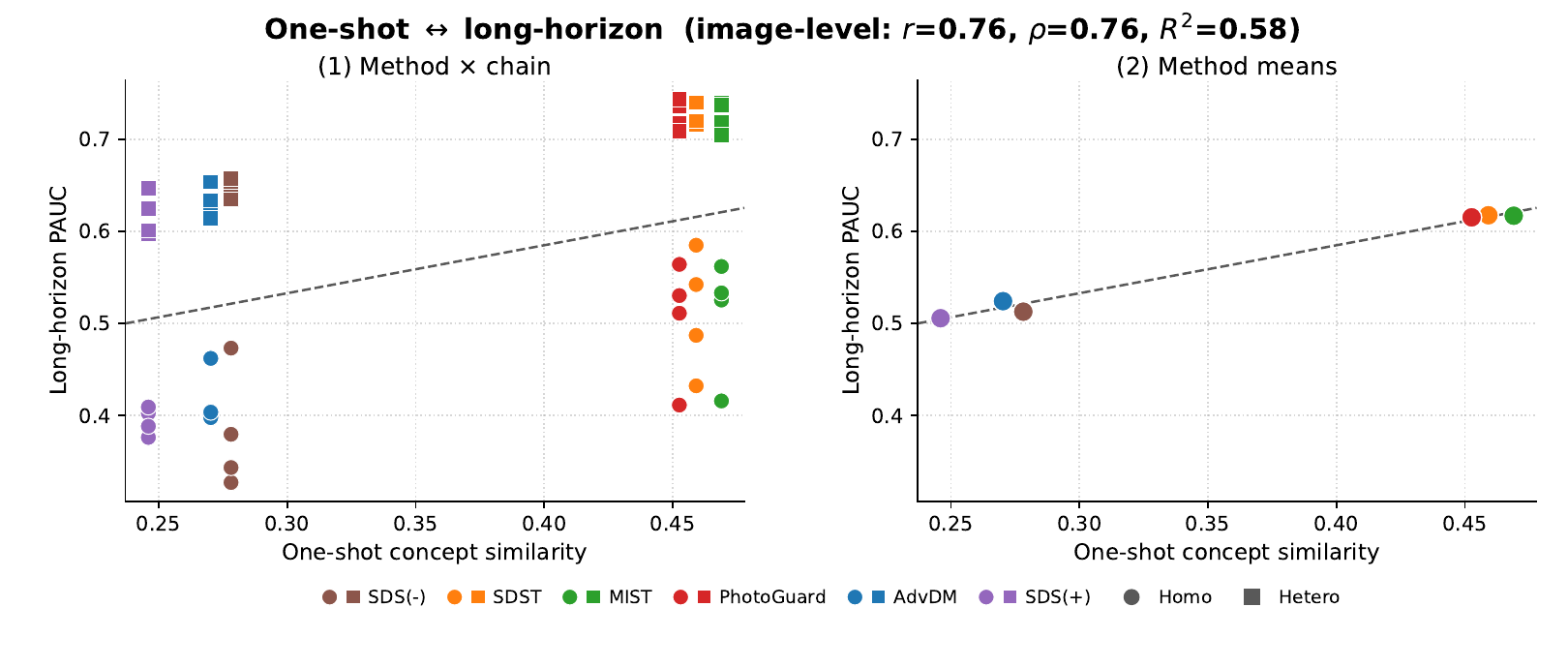}
\caption{One-shot semantic similarity versus long-horizon PAUC.
Left: method$\times$chain points (circles: homogeneous; squares: heterogeneous), used as the primary regression evidence.
Right: method-level means.
The dashed line is the OLS fit.
Title statistics correspond to image-level pooling and are stronger than the method$\times$chain $R^{2}$ reported in the main text.}
\label{fig:oneshot_vs_pauc}
\end{figure}

\begin{figure}[t]
\centering
\includegraphics[width=\linewidth]{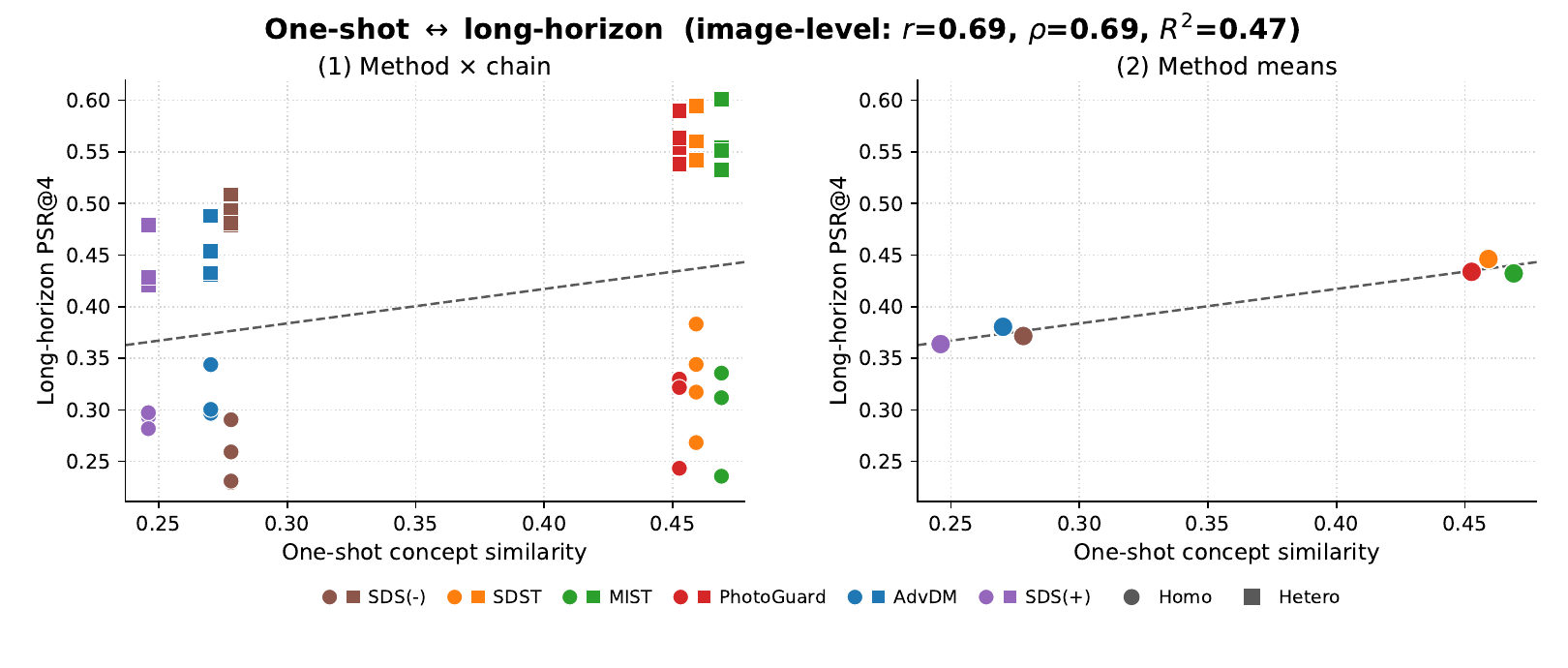}
\caption{Same protocol as Fig.~\ref{fig:oneshot_vs_pauc}, with long-horizon PSR at depth~$4$ as the vertical axis.}
\label{fig:oneshot_vs_psr4}
\end{figure}

\noindent
An analogous scatter plot with PSR@$4$ as the target
(Fig.~\ref{fig:oneshot_vs_psr4}) shows the same qualitative conclusion.
Together, these results support the main-text claim that endpoint one-shot measurements are insufficient to characterize protection under multi-step editing.

\subsection{Complementary Evaluation Under Protection Removal Attacks}

As discussed in the main paper, PREEMPT targets a realistic distribution and editing lifecycle after release.
Purification answers a different question: whether an informed adversary can strip protective signals before further use.
We therefore report complementary purification results here and do not include them in the main evaluation protocol.

\paragraph{Setup.}
We evaluate two representative purification attacks on face images protected by AdvDM, PhotoGuard, and Mist ($n{=}30$ images per method).
(1)~\textbf{DiffPure}~\cite{nie2022DiffPure}: guided diffusion purification with noise level $t{=}400$ (official default schedule with 1000 diffusion steps).
(2)\textbf{IMPRESS}\cite{cao2023impress}: 1000 optimization iterations with $\varepsilon{=}0.1$, learning rate $0.005$, $\alpha{=}0.01$, and noise $0.05$, following the authors' recommended hyperparameters.

\paragraph{Metric.}
Let $S_{\mathrm{id}}(\cdot,\cdot)$ denote ArcFace identity similarity.
For each protected image $I_{\mathrm{prot}}$ and its purified counterpart $I_{\mathrm{pur}}$, with clean original $I_{\mathrm{clean}}$, we report
\begin{equation}
\Delta_{\mathrm{id}}
=
S_{\mathrm{id}}(I_{\mathrm{clean}}, I_{\mathrm{pur}})
-
S_{\mathrm{id}}(I_{\mathrm{clean}}, I_{\mathrm{prot}}).
\end{equation}
Negative $\Delta_{\mathrm{id}}$ means purification moves the image farther from the clean identity than the protected image already was.
This analysis is distinct from the clean trajectory control $\Delta$ used elsewhere in the paper.

\paragraph{Results.}
Table~\ref{tab:purification} summarizes the results.
DiffPure strongly alters protected faces (mean $\Delta_{\mathrm{id}}{\approx}{-}0.72$), while IMPRESS induces only small identity shifts (mean $\Delta_{\mathrm{id}}{\approx}{-}0.03$).
These results characterize targeted removal rather than post release editing robustness, and therefore remain outside the main PREEMPT protocol.

\begin{table*}[t]
\centering
\caption{Purification on protected faces.
Metric: ArcFace identity similarity and
$\Delta_{\mathrm{id}}{=}S(\mathrm{clean},\mathrm{pur}){-}S(\mathrm{clean},\mathrm{prot})$.
DiffPure uses $t{=}400$; IMPRESS uses 1000 iterations with the authors' recommended hyperparameters.}
\label{tab:purification}
\begin{tabular}{llcccc}
\toprule
Attack & Method & $n$ & $S(\mathrm{clean},\mathrm{prot})$ & $S(\mathrm{clean},\mathrm{pur})$ & $\Delta_{\mathrm{id}}$ \\
\midrule
DiffPure & AdvDM & 30 & 0.933 & 0.230 & $-0.703$ \\
DiffPure & Mist & 30 & 0.910 & 0.174 & $-0.736$ \\
DiffPure & PhotoGuard & 30 & 0.906 & 0.184 & $-0.721$ \\
DiffPure & Mean & 90 & 0.916 & 0.196 & $-0.720$ \\
\midrule
IMPRESS & AdvDM & 30 & 0.933 & 0.924 & $-0.009$ \\
IMPRESS & Mist & 30 & 0.910 & 0.868 & $-0.042$ \\
IMPRESS & PhotoGuard & 30 & 0.906 & 0.862 & $-0.044$ \\
IMPRESS & Mean & 90 & 0.916 & 0.884 & $-0.032$ \\
\bottomrule
\end{tabular}
\end{table*}
{
    \small
    \bibliographystyle{ieeenat_fullname}
    \bibliography{main}
}